\documentclass[lettersize,journal]{IEEEtran}
\usepackage{amsmath,amsfonts}
\usepackage{algorithmic}
\usepackage{algorithm}
\usepackage{array}
\usepackage[caption=false,font=normalsize,labelfont=sf,textfont=sf]{subfig}
\usepackage{textcomp}
\usepackage{stfloats}
\usepackage{url}
\usepackage{verbatim}
\usepackage{graphicx}
\usepackage{cite}
\usepackage{xcolor}
\usepackage{booktabs}
\usepackage{booktabs, tabularx, multirow}
\newcolumntype{Y}{>{\raggedright\arraybackslash}X}

\begin{document}

\title{
From Freehand Demonstrations to Autonomous Robotic Ultrasound: An Anatomy-Anchored Learning Framework
}

\author{Yuan Bi\IEEEauthorrefmark{1}, Fangxu Han\IEEEauthorrefmark{1}, Ye Su, Yiping Zhou, Zhongliang Jiang and Nassir Navab%

\thanks{Manuscript received Month DD, YYYY; revised Month DD, YYYY. (Corresponding author: Zhongliang Jiang.)}
\thanks{\IEEEauthorrefmark{1}These authors contributed equally to this work.}%
\thanks{Yuan Bi, Fangxu Han, Ye Su, Yiping Zhou and Nassir Navab are with the Chair for Computer Aided Medical Procedures (CAMP), Technical University of Munich, Munich, Germany.}%
\thanks{Yuan Bi is with the Munich Center for Machine Learning (MCML), Munich, Germany.}
\thanks{Zhongliang Jiang is with the Department of Mechanical Engineering, The University of Hong Kong, Hong Kong SAR, China.}
}


\markboth{Journal of \LaTeX\ Class Files,~Vol.~14, No.~8, August~2021}%
{Shell \MakeLowercase{\textit{et al.}}: A Sample Article Using IEEEtran.cls for IEEE Journals}


\maketitle

\begin{abstract}
Ultrasound imaging is highly operator-dependent, motivating robotic systems that can reproduce expert scanning skills consistently. However, these skills involve coordinated interpretation of anatomical landmarks, ultrasound images, and probe-tissue interactions that are difficult to encode in manually designed control strategies. Imitation learning offers a promising alternative, but its observation and action representations should reflect the anatomically structured nature of scanning. We propose an anatomy-anchored imitation-learning framework that integrates anatomy-guided probe placement with learned ultrasound navigation. The placement module identifies a task-specific anatomical starting point from external visual observations and establishes an initial probe pose as the reference frame for subsequent scanning. A flow-based policy combines anchor-relative probe poses, ultrasound images, probe-mounted camera observations, and contact forces to generate scanning motions, while a termination head predicts task completion. This formulation avoids continuous reliance on third-person vision and reduces dependence on the robot-patient configuration. A handheld acquisition platform enables multimodal demonstrations to be collected during natural freehand scanning without a robot. On 16 previously unseen participants, task success rates reach 96\%, 94\%, 88\%, and 94\% for kidney, gallbladder, and liver standard-plane acquisition and radial artery sweeping, respectively. Compared with a fine-tuned $\pi_0$ model, success on the abdominal tasks improves by 10-42 percentage points, with reduced time-to-target, peak contact force, and trajectory jerk. Ablation studies support the complementary contributions of anatomical anchoring, egocentric vision, and force feedback. End-to-end trials with automatic probe placement achieve 90-100\% success, and the policies transfer without retraining to a second robot with 83-100\% success on the selected tasks. Code and videos are available at \url{https://anonymous.4open.science/r/Anatomy_anchored_SFP-5B0E/}

\end{abstract}



\section{Introduction}
Ultrasound is one of the most widely used medical imaging modalities owing to its real-time imaging capability, non-invasive nature, portability, and relatively low cost~\cite{jiang2023robotic}. It plays an important role in a broad range of diagnostic and interventional procedures. However, conventional ultrasound scanning remains highly operator-dependent. This dependence on individual skill can lead to considerable variability in image quality and examination outcomes, limiting the standardization and accessibility of ultrasound imaging~\cite{li2021overview}. Robotic ultrasound offers a promising solution to these challenges by enabling precise, repeatable, and consistent probe manipulation~\cite{bi2024machine}. Nevertheless, autonomous ultrasound scanning is difficult to achieve using manually designed control strategies alone. Many scanning skills rely on subtle adjustments learned through extensive clinical experience. Such expert behaviors are often difficult to formulate explicitly as predefined trajectories or control laws~\cite{jiang2024intelligent}. Robot learning, and imitation learning in particular, therefore provides a promising framework for acquiring ultrasound scanning skills directly from expert demonstrations and transferring them to autonomous robotic systems.

\begin{figure}
    \centering
    \includegraphics[width=0.9\linewidth]{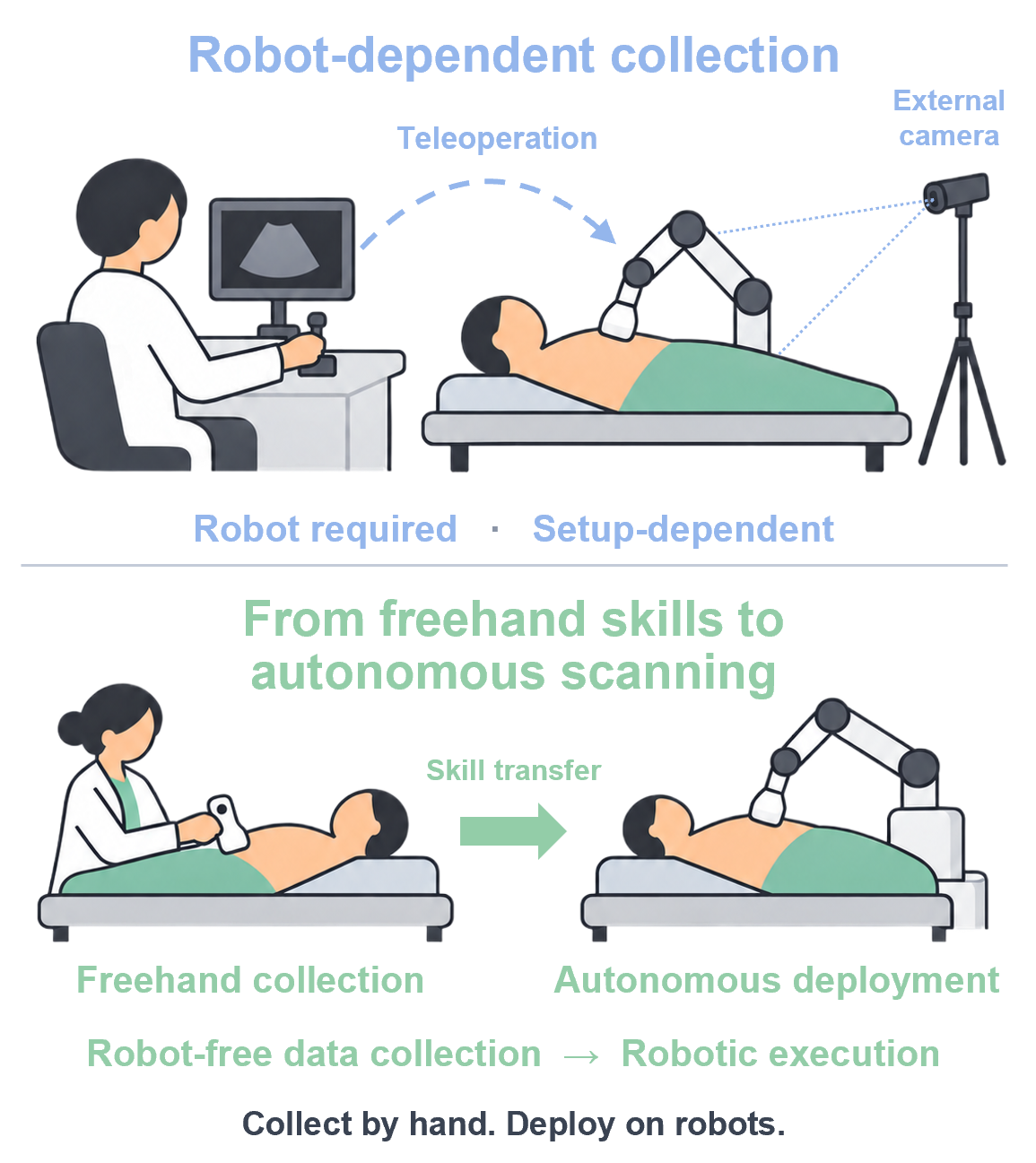}
    \caption{From freehand demonstrations to autonomous robotic ultrasound. Top: Robot-dependent demonstration collection requires teleoperation and a dedicated robotic setup. Bottom: Our framework enables robot-free collection of expert scanning demonstrations and transfers the acquired skills to autonomous robotic execution through anatomy-anchored imitation learning.}
    \label{fig:teaser}
\end{figure}

Imitation learning enables robots to acquire control policies directly from expert demonstrations by learning a mapping from sensory observations to robot actions~\cite{zare2024survey}. Early approaches commonly formulated this problem as behavior cloning~\cite{torabi2018behavioral}, in which a policy is trained to reproduce the actions demonstrated under corresponding observations. More recently, generative visuomotor policies have substantially improved the ability to model complex and multimodal robot behaviors, such as Diffusion Policy~\cite{chi2025diffusion} or Flow-based policies~\cite{jiang2025streaming,zhang2025flowpolicy}. In parallel, vision-language-action models have incorporated pretrained vision-language representations and large-scale robot datasets to improve semantic reasoning and generalization across manipulation tasks~\cite{black2024pi0,physicalintelligence2025pi05}.

Despite these advances in policy architectures, most imitation-learning frameworks are still designed around the characteristics of general robotic manipulation. Their observations commonly include one or more third-person or wrist-mounted camera views together with robot states, such as joint configurations or end-effector poses~\cite{khazatsky2024droid}, while force or tactile measurements may additionally be used for contact-rich tasks~\cite{zhang2025ta}. Such observations are well suited to manipulation, where the identities, positions, and spatial arrangements of objects can vary substantially across demonstrations and deployment environments. External vision provides the global scene context required to interpret these variations, while robot states are often represented in a global or robot-base coordinate frame to relate the end effector to the surrounding workspace~\cite{o2024open}.

Imitation-learning methods for robotic ultrasound, however, adopt different observation configurations. Some retain the multimodal design of general manipulation by combining third-person and wrist-mounted camera views with ultrasound images and robot states~\cite{nelson2026open}, whereas others rely primarily on ultrasound images and probe pose~\cite{jiang2025towards}. Despite this diversity, the necessity and task-specific role of these modalities have not been systematically examined. This motivates us to revisit the observation design for robotic ultrasound and identify the sensory inputs and pose representations that best reflect the structure of the scanning task.

Unlike general manipulation, robotic ultrasound is an anatomy-conditioned and locally constrained interaction once the probe has been initialized at the target anatomical region. At this stage, continuous third-person observations may provide limited additional information for probe navigation. They also introduce dependence on the workspace configuration. These setup-specific dependencies complicate demonstration collection and can limit generalization across environments and robotic platforms. We therefore argue that external vision is better used for anatomical localization and initial probe placement, rather than as a continuous observation during scanning.

However, removing third-person vision does not imply that ultrasound images alone provide a sufficient representation of the scanning process. Unlike conventional RGB videos, ultrasound images do not directly depict the motion of the probe relative to the patient. Each frame represents a two-dimensional cross-section of three-dimensional anatomy, and out-of-plane probe motion cannot be reliably tracked based only on ultrasound images. Ultrasound feedback alone therefore provides an incomplete and potentially ambiguous representation of probe motion and spatial context.

These observations motivate an anatomy-anchored egocentric formulation for robot learning in ultrasound. We decompose autonomous ultrasound scanning into two phases: an anatomy-guided, rule-based probe initialization phase and a learned probe navigation phase. During initialization, anatomical feature detection is used to place the probe at a scan-specific starting pose. Once stable contact has been established, the learned policy takes over and performs local probe navigation using four complementary modalities: ultrasound images, a probe-mounted wrist-camera view, contact force, and probe pose relative to an anatomical anchor. The ultrasound image provides direct information about the internal anatomy, while the wrist camera captures the local interaction between the probe and the scanning region, including probe motion relative to the patient surface. Force measurements characterize the contact condition, which strongly affects both image quality and tissue deformation.

Because ultrasound scanning is organized around anatomical structures and scan-specific starting locations, the global pose of the probe in the robot-base frame is of limited relevance to the underlying scanning skill. Although anatomy varies across patients, the relative organization of anatomical structures is consistent, and clinical scans are typically initialized from identifiable landmarks or predefined body regions. We therefore use the initial probe pose, established relative to a detected anatomical feature, as the reference frame for the subsequent scan. Expressing probe motion relative to this anatomically anchored initial frame removes variations caused by robot placement, patient positioning, and workspace configuration while preserving information about local scan progression.

This probe-centric observation design also enables demonstrations to be collected independently of the robotic platform. We develop a dedicated handheld demonstration platform that synchronously records ultrasound images, wrist-camera views, probe pose, and contact force while an expert performs natural manual scanning. This eliminates the need to collect demonstrations by remotely controlling the robot or physically guiding its arm. As a result, demonstrations can be acquired more flexibly and efficiently without being constrained by robot availability, workspace limitations, or a specific robotic embodiment.

\begin{figure*}
    \centering
    \includegraphics[width=0.9\linewidth]{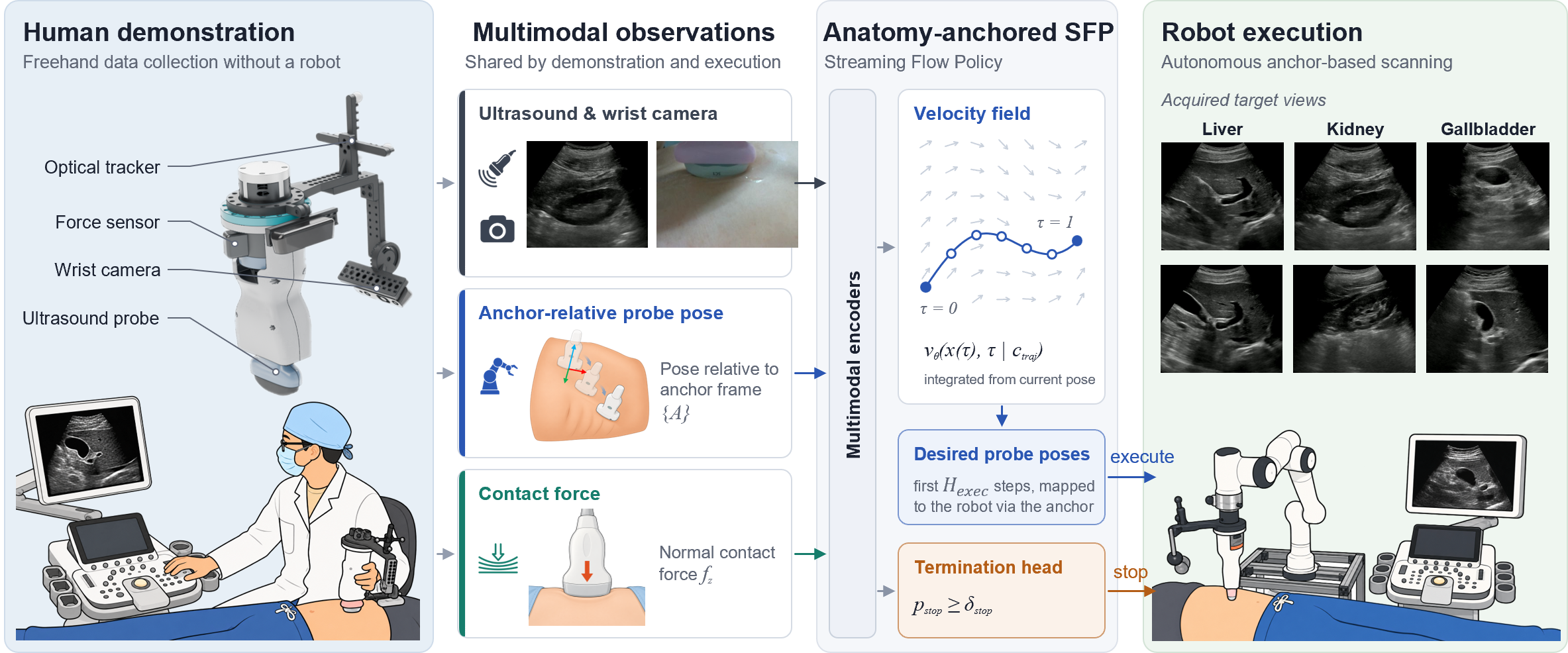}
    \caption{Overview of the proposed anatomy-anchored imitation-learning framework. A handheld acquisition platform records synchronized probe poses, contact forces, ultrasound images, and wrist-camera views during freehand demonstrations. These multimodal observations, together with an anchor-relative pose representation, enable learning scanning policies for autonomous robotic ultrasound.}
    \label{fig:overview}
\end{figure*}

The contributions are summarized as follow:


  

\begin{itemize}
  \item We propose an anatomy-anchored imitation-learning framework for robotic ultrasound that separates scanning into anatomy-guided probe placement and learned navigation. A Streaming Flow Policy generates probe motion relative to an anatomical anchor frame from ultrasound images, probe-mounted camera views, and contact force, while a dedicated termination head decides when the target view has been reached. 

  \item We develop a handheld multimodal acquisition platform that synchronously records ultrasound images, wrist-camera views, probe pose, and contact force during natural freehand scanning. Together with the anchor-relative formulation, it allows demonstrations to be collected without a robot and the learned policies to be deployed on different robotic platforms.

  \item We validate the framework on 16 previously unseen participants across kidney, gallbladder, and liver standard-plane acquisition and radial artery sweeping, achieving $88$-$96\%$ task success and outperforming a fine-tuned $\pi_0$ model by $10$-$42$ percentage points on the abdominal tasks. Ablations show the complementary roles of wrist-camera observation, force feedback, and anchor-relative poses, and additional experiments demonstrate end-to-end scanning with automatic probe placement and its ability to transfer to a second robot without retraining.
\end{itemize}



\section{Related Work}
\subsection{Imitation Learning in General Domain}
Imitation learning has been widely used to acquire manipulation skills directly from expert demonstrations. Early behavior cloning methods learned observation-to-action mappings, while recurrent models such as LSTM-based BC-RNN incorporated observation history to capture temporal dependencies and the non-Markovian nature of human demonstrations. Mandlekar~\emph{et al.} demonstrated the benefits of such history-dependent policies, particularly for long-horizon manipulation tasks~\cite{mandlekar2021matters}. With the rise of Transformers, Behavior Transformer introduced action discretization with residual prediction to model multimodal behaviors~\cite{shafiullah2022behavior}, while Action Chunking with Transformers predicts sequences of future actions to reduce the effective task horizon and mitigate compounding errors~\cite{zhao2023learning}. More recently, MaIL explored Mamba-based selective state-space models as an efficient alternative to Transformer backbones, particularly under limited demonstration data~\cite{jia2024mail}.

Beyond sequence-modeling architectures, generative policies have substantially improved the modeling of multimodal robot behaviors. Diffusion Policy generates action sequences through iterative denoising, providing expressive multimodal prediction and temporally consistent control, but at the cost of multiple inference steps~\cite{chi2025diffusion}. Flow-based approaches aim to reduce this overhead: AdaFlow adaptively allocates integration steps according to action uncertainty~\cite{hu2024adaflow}, while FlowPolicy uses consistency flow matching to enable single-step action generation~\cite{zhang2025flowpolicy}.

Recent work further targets tighter and faster sensorimotor loops. Streaming Diffusion Policy reuses partially denoised action trajectories across control steps to accelerate diffusion-based inference~\cite{hoeg2025fast}. Streaming Flow Policy goes further by treating the action trajectory itself as the flow trajectory, allowing actions to be generated and executed incrementally rather than waiting for an entire action chunk to be synthesized~\cite{jiang2025streaming}. This progression is particularly relevant to robotic ultrasound, where effective imitation learning requires modeling temporally coordinated probe motions from multimodal sensory observations while supporting fast and responsive closed-loop execution.

\subsection{Handheld Demonstration Collection Platform}
Collecting demonstrations directly through robot teleoperation can be time-consuming and restricts data acquisition to environments where the robot is available. Recent work in robot manipulation has therefore explored portable interfaces that decouple demonstration collection from the physical robot, allowing humans to provide demonstrations more naturally and in diverse environments. The Universal Manipulation Interface (UMI) introduced a handheld gripper equipped with onboard sensing to collect in-the-wild manipulation demonstrations without requiring a robot during data acquisition~\cite{chi2024universal}. Together with a relative-trajectory action representation, demonstrations collected with UMI can be transferred to different robot platforms. In parallel, DexCap proposed a portable motion-capture system that records natural human wrist and finger motions together with 3D observations, which are subsequently retargeted to dexterous robot hands~\cite{wang2024dexcap}. These approaches demonstrate two complementary strategies for robot-free data collection: using a robot-like handheld interface to reduce the embodiment gap, or directly capturing natural human motion and addressing this gap through retargeting.

Subsequent works have extended portable demonstration interfaces with richer sensing and feedback. ManiWAV augmented a UMI-style handheld interface with contact audio, enabling the collection of synchronized audio-visual demonstrations for contact-rich manipulation~\cite{liu2024maniwav}. ARCap instead addressed the feasibility of robot-free demonstrations by providing augmented-reality visualization and haptic warnings, guiding users to generate motions that respect the kinematic and collision constraints of the target robot~\cite{chen2025arcap}. More recent approaches have increasingly incorporated interaction sensing. FreeTacMan introduced wearable visuo-tactile grippers for collecting visual, tactile, and motion information directly from human manipulation~\cite{wu2025freetacman}, while UMI-FT integrated six-axis force/torque sensing into a handheld interface to acquire RGB, depth, pose, and interaction wrench for learning compliant manipulation policies~\cite{choi2026wild}. Collectively, these works highlight a shift from robot-dependent teleoperation toward portable, multimodal demonstration systems that can capture human manipulation skills without requiring the target robot during data collection.

This paradigm is particularly attractive for robotic ultrasound, where demonstrations can be acquired directly with the same physical instrument used during robotic execution. Rather than constructing a surrogate end-effector or retargeting human hand motions to a robot gripper, a sonographer can manipulate an instrumented ultrasound probe while synchronized ultrasound images, local visual observations, contact force, and probe motion are recorded. This substantially reduces the human-to-robot embodiment gap while retaining the flexibility and scalability of robot-free demonstration collection.

\subsection{Robot Learning for Robotic Ultrasound}

Reinforcement learning has been explored for autonomous ultrasound navigation~\cite{hase2020ultrasound,li2021autonomous,ning2021autonomic}, but its reliance on trial-and-error training makes direct learning on patients impractical. Training in simulated or simplified environments is therefore common, yet the limited realism of ultrasound simulation creates a substantial sim-to-real gap~\cite{bi2022vesnet}. Imitation learning offers a more practical alternative by learning scanning strategies directly from expert demonstrations.

Early movement-primitive methods focused mainly on reproducing demonstrated scanning patterns, with limited online action reasoning. Dall’Alba \emph{et al.}~\cite{dall2024imitation} mapped the probe’s relative position along a predefined trajectory to a learned compression-force profile, without ultrasound-image feedback.

Other approaches inferred probe actions directly from ultrasound images, without including the robot pose in the policy input. Wang \emph{et al.}~\cite{wang2024robotic} used two consecutive ultrasound frames to predict translational and rotational increments for standard-view acquisition, while contact force was handled separately by an admittance controller. Similarly, UltraBot~\cite{jiang2025towards} mapped the current ultrasound image to one of several discrete 6-DoF probe motions or a stage-transition command. Although both methods are image-only at inference time, their action labels are still defined in the robot or scanner base coordinate system, rather than in an anatomy-anchored frame.

A further group of methods conditioned the policy on both ultrasound observations and robot pose. Z. Ma \emph{et al.}~\cite{ma2026hardware} collected in vivo kidney-scanning demonstrations through teleoperation and conditioned an Action Chunking Transformer on the ultrasound image and current Cartesian end-effector pose. The policy predicts sparse task-space waypoints rather than incremental motions to reduce long-horizon error accumulation and support transfer across robot instances. However, the pose representation is not explicitly anchored to the patient anatomy. In contrast, J. Ma \emph{et al.}~\cite{ma2026autonomous} collected demonstrations by hand-guiding the robot and expressed probe trajectories in an anatomical frame defined by the navel and nipple locations, combined with a segmentation-derived ultrasound state to infer the next probe pose. This frame, however, must be reconstructed from RGB-D observations whenever the subject's position or posture changes, and the method was validated only on an abdominal phantom. Our anchor frame is instead defined by the probe-tip pose at initial contact with a single landmark and is established once per episode without frame reconstruction, which enables freehand demonstration collection and evaluation on human participants.

Beyond ultrasound and proprioception, multimodal approaches have incorporated visual and force information to represent probe–patient interaction more comprehensively. Si \emph{et al.}~\cite{si2023unified} collected demonstrations through teleoperation and fused ultrasound images, external RGB images, contact force, and end-effector pose in the robot-base frame to predict desired probe poses and forces. Their later journal extension~\cite{si2025deep} incorporated trajectory optimization and compliant control and expanded the evaluation from phantom experiments to human scanning. UltraDP~\cite{chen2025ultradp} instead collected demonstrations through direct physical guidance, with sonographers manually moving the robot by holding its flange during scanning. Its diffusion policy uses ultrasound images, wrist-camera RGB-D images, contact wrench, and probe pose in the robot-base frame. Its ultrasound processing is tailored to carotid scanning, using a pretrained encoder and guidance module to detect and center the artery, which may limit transfer to other anatomical targets.


Overall, previous advances provide promising foundations for learning ultrasound scanning skills, but their application requires coordinated choices in policy learning, data collection, and observation representation. Policies should generate smooth, continuous probe trajectories while supporting timely updates from imaging and contact feedback. Portable acquisition interfaces can capture these complementary signals during natural freehand scanning, but many existing ultrasound-learning systems still rely on robot teleoperation or physical guidance of the robot arm for demonstration collection. Meanwhile, observation designs range from ultrasound images alone to extensive multimodal inputs, offering different levels of information about anatomy, probe motion, and contact conditions. Robot-base pose representations can introduce dependence on the workspace configuration, whereas anatomical representations may require additional reconstruction and calibration. These considerations motivate an integrated framework that combines smooth, responsive trajectory generation, robot-independent multimodal demonstration collection, and anatomically grounded observations that remain practical to acquire and use during both demonstration and deployment.


\section{Methods}

Imitation learning for robotic ultrasound should ultimately aim to reproduce not only the probe trajectories demonstrated by human operators, but also the perceptual cues and spatial reasoning that underlie their scanning decisions. Experienced sonographers rarely begin a scan from an arbitrary probe pose. Instead, depending on the examination, they typically place the probe at a recognizable anatomical landmark before starting ultrasound-guided navigation. Such landmarks are usually surface-visible anatomical features that can be identified directly from the patient's external appearance, rather than structures that must first be localized in the ultrasound image. The initial probe placement is therefore primarily guided by visual observation of the patient surface. Once the probe is positioned at this anatomical starting point, the subsequent scanning motion is performed relative to this reference. Although the choice of landmark varies across anatomical regions and scanning protocols, it remains relatively consistent for acquiring the same standard plane.

\begin{figure}[H]
    \centering
    \includegraphics[width=\linewidth]{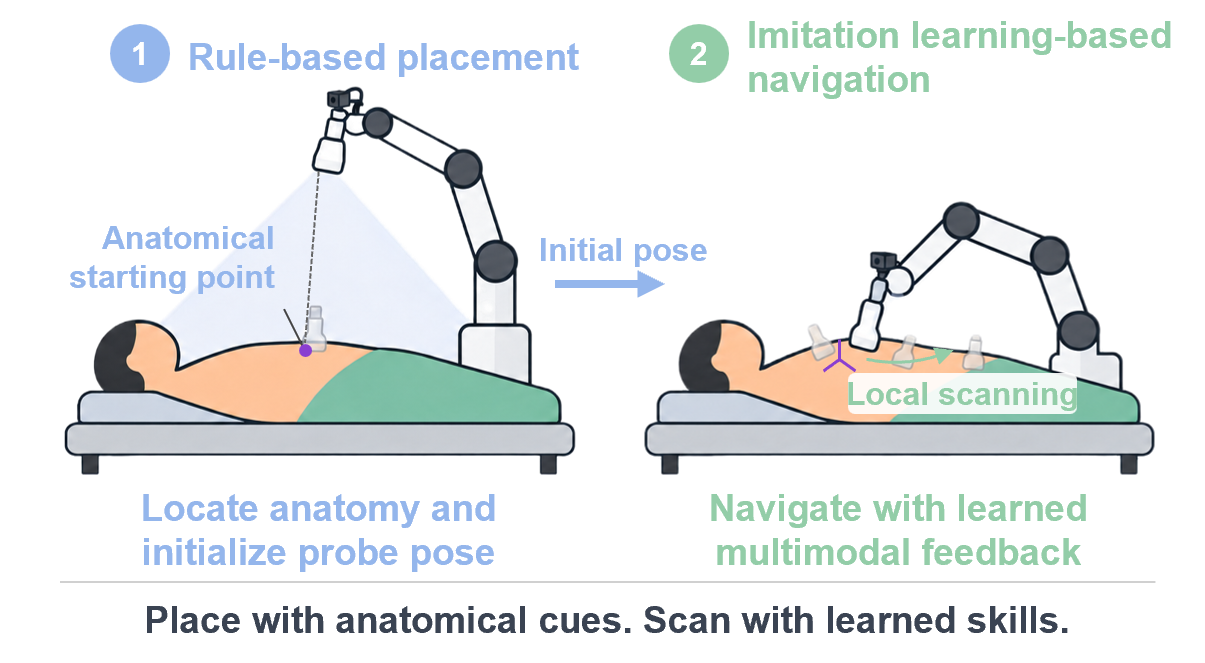}
    \caption{Separation of probe placement and ultrasound navigation. Rule-based placement uses external anatomical cues to establish the initial probe pose and anatomical anchor. The learned policy then performs local scanning relative to this anchor using multimodal feedback.}
    \label{fig:two_step}
\end{figure}

This observation naturally decomposes the overall scanning procedure into two stages: probe placement and ultrasound-guided navigation. The placement stage establishes an anatomy-dependent starting pose from an externally observable landmark, while the navigation stage searches for the desired imaging plane relative to this anatomical anchor. Accordingly, probe motion during navigation can be represented in an anatomy-anchored coordinate system rather than solely in the absolute robot-base frame. During this stage, the ultrasound image provides the primary source of information for determining how the probe should move, while occasional visual observation of the local body surface provides complementary context about the probe and the surrounding scanning region. Meanwhile, contact force is continuously perceived through tactile feedback and used to regulate probe–patient interaction. Motivated by these observations, we formulate robotic ultrasound imitation learning around four complementary sources of information: anatomy-anchored probe motion, ultrasound images, a probe-mounted wrist-camera view, and contact force.

This formulation also enables demonstrations to be collected using a portable freehand acquisition platform rather than directly on a robotic system. By instrumenting a handheld ultrasound probe with the required sensing modalities, expert demonstrations can be recorded during natural freehand scanning without requiring a robot in the data-collection loop. This substantially reduces the hardware requirements for demonstration collection and makes it easier to acquire data across different anatomical targets, scanning environments, and operators.

To further promote smooth and temporally consistent probe motion, we adopt a Streaming Flow Policy defined over the anatomy-anchored action space. In contrast to diffusion-based policies that iteratively generate action trajectories, the flow-based formulation learns a continuous velocity field that transports actions toward the demonstrated motion distribution. This formulation is particularly compatible with our anatomy-anchored representation, where probe motion is expressed as continuous changes relative to a meaningful anatomical reference. In the following, we first describe the handheld demonstration-acquisition platform, then present the anatomy-anchored imitation-learning formulation and Streaming Flow Policy for ultrasound navigation, and finally introduce the anatomy-guided probe placement module.

\subsection{Data Acquisition Platform}

The design of the Portable Ultrasound Demonstration Acquisition Platform is guided by three key principles: \textbf{a) Freehand acquisition:} The acquisition device should allow sonographers to perform scans naturally in a freehand manner, without being constrained by a robotic embodiment during demonstration collection. \textbf{b) Sufficiency:} The acquired multimodal data should capture the information necessary for effectively learning ultrasound scanning behaviors, including both the perceptual observations and interaction states relevant to probe manipulation. \textbf{c) Seamless transfer:} The acquisition device should support straightforward integration with different robotic embodiments, enabling the collected demonstrations to be transferred to and executed on robotic systems with minimal adaptation.


\begin{figure}
    \centering
    \includegraphics[width=\linewidth]{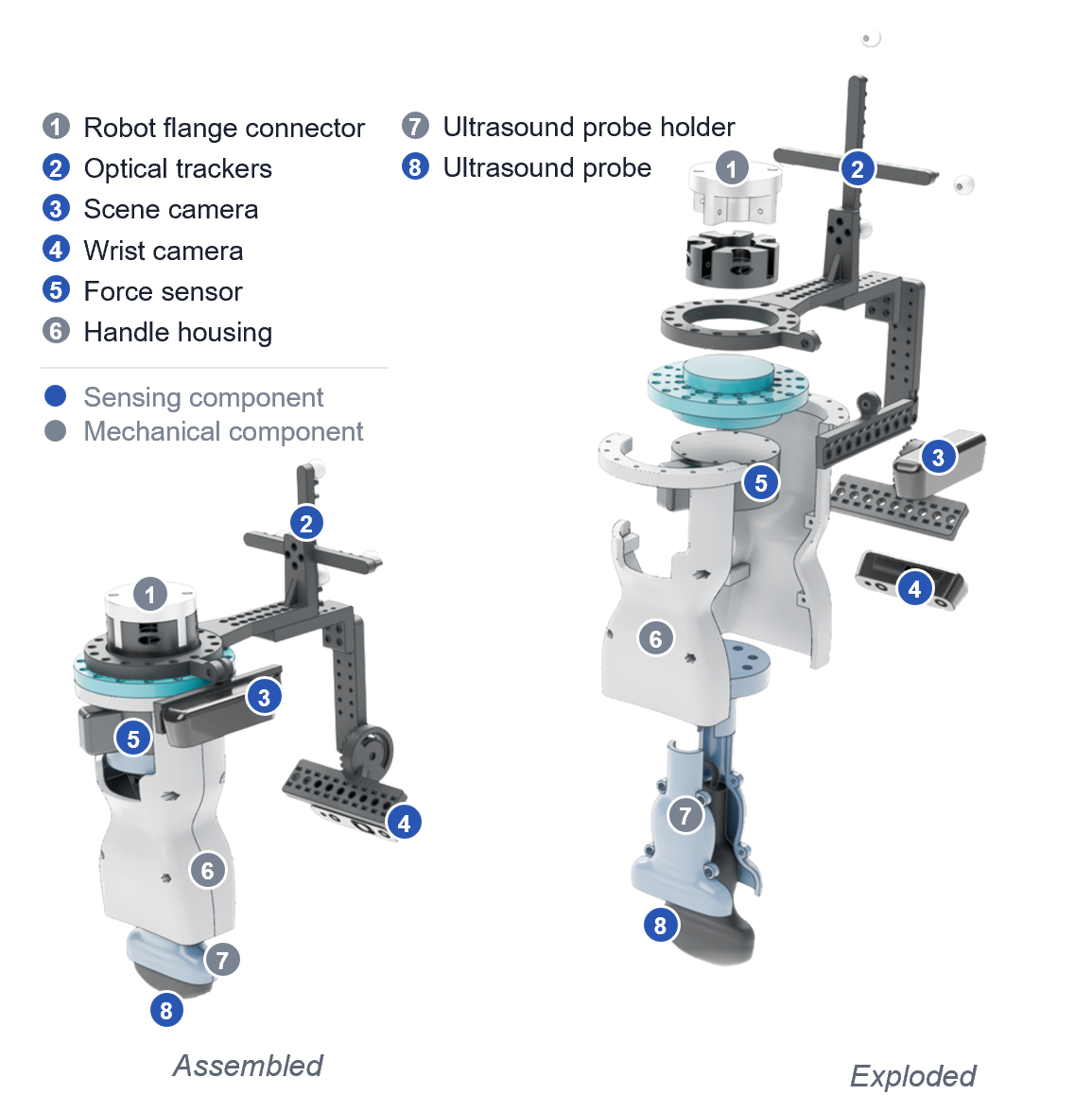}
    \caption{Portable ultrasound demonstration acquisition platform. Assembled (left) and exploded (right) views show the integrated sensing components, handheld housing, probe holder, and robot mounting interface, supporting multimodal freehand demonstration collection and subsequent robotic deployment.}
    \label{fig:acquisition_platform}
\end{figure}

\subsubsection{Sensors} Following these design principles, we developed the portable acquisition device shown in Fig.~\ref{fig:acquisition_platform}. The platform records four synchronized information streams: the ultrasound probe pose, applied force, wrist-camera view, and ultrasound images. The \textbf{probe pose} is tracked using an NDI optical tracking system, with a tracking marker rigidly mounted to the body of the acquisition device. A one-time calibration is performed to determine the rigid transformation between the tracking marker and the ultrasound probe tip, allowing the probe pose to be recovered throughout the scan. The \textbf{applied force} is measured using a force sensor mounted between the handheld shell and the ultrasound probe holder. The shell encloses the probe assembly and serves as the interface through which the sonographer holds and manipulates the device, such that the force transmitted from the operator to the scanning target is measured by the sensor. Importantly, the same mechanical load path is preserved when the device is attached to the robot end-effector, allowing the contact force to be measured consistently during both demonstration collection and robotic execution. The \textbf{wrist camera} is positioned to observe the probe footprint together with its local surrounding region on the patient surface, providing a probe-centric visual observation of the probe-patient interaction and local probe motion. An additional \textbf{scene camera} captures a broader view of the patient and the three-dimensional surface geometry to support anatomical localization and initial probe placement. Finally, the \textbf{ultrasound probe} is mounted using a dedicated probe holder, while the corresponding ultrasound images are acquired directly from the ultrasound system through a frame grabber.


\subsubsection{Calibration and Synchronization} Two calibration procedures are performed before data acquisition: spatial calibration between the NDI tracking marker and the ultrasound probe, and calibration of the force sensor.

For the \textbf{marker-to-probe calibration}, the acquisition device is first rigidly mounted to the robot end-effector. A hand-eye calibration is then performed to obtain ${}^{EE}\mathbf{T}_{M}$, which represents the transformation from the NDI marker frame $M$ to the robot end-effector frame $EE$. Since the geometric relationship between the robot mounting interface and the ultrasound probe is fixed by the mechanical design, ${}^{P}\mathbf{T}_{EE}$, the transformation from the end-effector frame $EE$ to the probe frame $P$, is obtained directly from the CAD model. The transformation from the NDI marker frame to the probe frame can then be computed as ${}^{P}\mathbf{T}_{M} = {}^{P}\mathbf{T}_{EE} \,{}^{EE}\mathbf{T}_{M}$.
Once calibrated, this fixed transformation allows the pose of the ultrasound probe to be recovered directly from the NDI marker measurements during freehand demonstration acquisition. 

For the \textbf{force-sensor calibration}, the gravitational load introduced by the ultrasound probe, probe holder, and other components mounted below the force sensor is compensated. By removing this intrinsic load from the sensor measurements, the resulting force signal reflects the interaction force applied by the operator to the scanning target rather than the weight of the acquisition device itself.

In addition to spatial and force calibration, the recorded sensing streams are temporally synchronized. During acquisition, ultrasound images, wrist-camera images, NDI marker poses, and force measurements are published independently through ROS and recorded into a single rosbag. Temporal alignment is performed offline using the ultrasound stream as the reference timeline. For each ultrasound frame, the nearest measurement from each of the remaining modalities is selected according to its timestamp within a predefined modality-specific tolerance. Short interruptions in NDI tracking are handled by interpolating the probe pose between the nearest valid measurements preceding and following the missing interval.


\subsection{Anatomy-anchored Streaming Flow Policy}

We formulate imitation-based ultrasound navigation as learning a conditional velocity field in an anchor-relative probe-pose space. At each policy update, a history of ultrasound images, wrist-camera observations, anchor-relative probe poses, and contact forces is used to condition a Streaming Flow Policy (SFP) \cite{jiang2025streaming}. 
Given the currently measured probe pose, the policy network predicts its instantaneous velocity over normalized horizon-progress trajectory time. 
The predicted velocity field is integrated to obtain a sequence of desired probe poses. During receding-horizon execution, only the first part of the generated trajectory is executed before the policy is queried again with updated observations. In parallel, a separate termination head determines whether the target ultrasound view has been reached.


\begin{figure*}
    \centering
    \includegraphics[width=\linewidth]{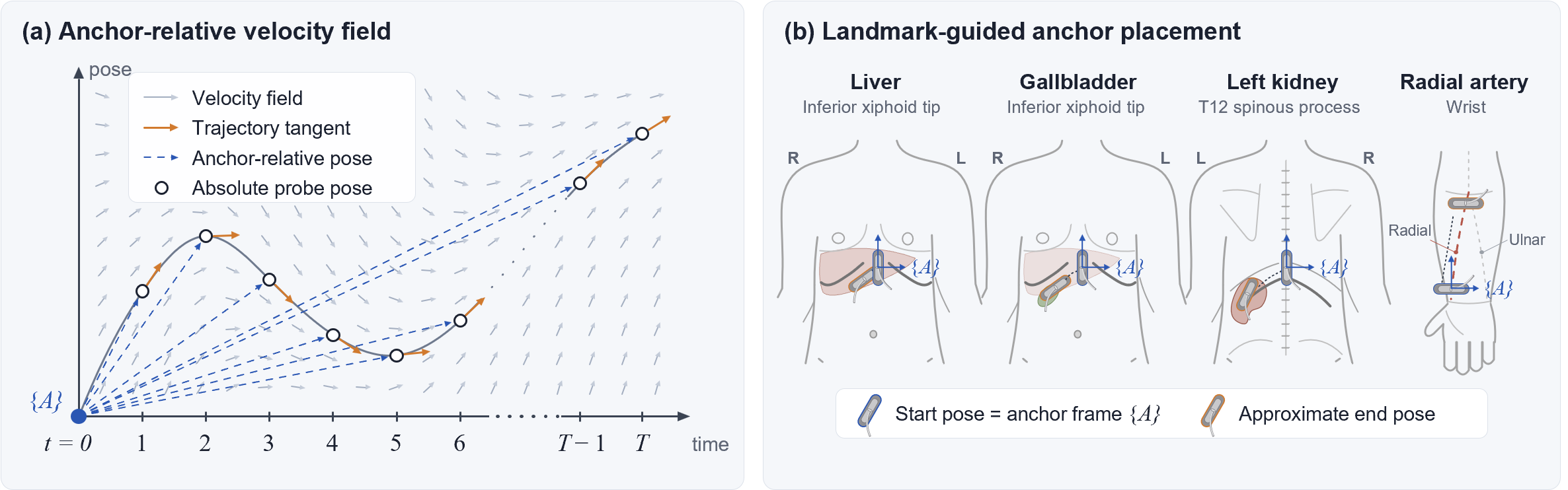}
    \caption{Anatomy-anchored formulation of ultrasound scanning. (a) Probe motion is modeled as a continuous velocity field in an anchor-relative pose space. (b) Landmark-guided probe placement establishes the initial anchor frame for subsequent ultrasound-guided navigation, illustrated for liver, gallbladder, left-kidney, and radial artery scanning.}
    \label{fig:anchor}
\end{figure*}

\subsubsection{Anchor-Relative Navigation Formulation}

Let $\mathcal{F}_{G}$ denote the global frame in which the probe pose is measured and let $\mathcal{F}_{P_t}$ denote the probe frame at time step $t$. The definition of $\mathcal{F}_{G}$ depends on the embodiment: it corresponds to the optical tracker frame during freehand demonstration acquisition and to the robot-base frame during robotic execution.

At the beginning of each navigation episode, the initial probe pose is used to define an episode-specific anchor frame $\mathcal{F}_{A}$. Its pose in the corresponding global frame is ${}^{G}\mathbf{T}_{A}:={}^{G}\mathbf{T}_{P_0}$.
The anchor remains fixed throughout the episode. Each subsequent probe pose is expressed relative to this anchor as ${}^{A}\mathbf{T}_{P_t}=\left({}^{G}\mathbf{T}_{A}\right)^{-1}{}^{G}\mathbf{T}_{P_t}$.
Because the global frame is eliminated by this transformation, the resulting relative pose does not depend on whether the original probe trajectory was measured in the optical tracker frame or the robot-base frame. This provides a common pose representation for freehand demonstrations and robotic roll-outs.
Unlike the relative-trajectory representation of UMI~\cite{chi2024universal}, which expresses actions relative to the current end-effector pose, the anchor remains fixed throughout the episode, so the anchor-relative pose also encodes the progress of the scan with respect to the anatomical starting point.

We encode ${}^{A}\mathbf{T}_{P_t}$ as a 9-dimensional state vector

\begin{equation}
\mathbf{x}^{A}_{t}
=
\begin{bmatrix}
\mathbf{p}^{A}_{t} \\
\boldsymbol{\rho}
\left(
\mathbf{R}^{A}_{t}
\right)
\end{bmatrix}
\in \mathbb{R}^{9},
\label{eq:anchor_relative_state}
\end{equation}

where $\mathbf{p}^{A}_{t}\in\mathbb{R}^{3}$ is the relative translation and $\boldsymbol{\rho}(\mathbf{R}^{A}_{t})\in\mathbb{R}^{6}$ is the continuous 6D representation of the relative orientation~\cite{zhou2019continuity}.

At a policy update performed at time step $t$, the corresponding demonstrated trajectory over a prediction horizon of $H$ steps is represented as

\begin{equation}
\boldsymbol{\xi}_{t}
=
\left(
\mathbf{x}^{A}_{t},
\mathbf{x}^{A}_{t+1},
\ldots,
\mathbf{x}^{A}_{t+H}
\right).
\label{eq:anchor_relative_trajectory}
\end{equation}

The first element of $\boldsymbol{\xi}_{t}$ corresponds to the current probe state, while the remaining elements describe its future evolution relative to the fixed episode anchor. The policy does not directly regress these future poses. Instead, the anchor-relative poses define the state space in which the policy learns a conditional velocity field. During inference, this velocity field is initialized from the current anchor-relative probe state and integrated to generate the future desired probe poses.

Fig.~\ref{fig:anchor} illustrates the spatial and anatomical interpretation of this formulation. In Fig.~\ref{fig:anchor}(a), successive probe poses are expressed relative to the fixed initial anchor, while the trajectory tangents describe their continuous evolution along the scan. Fig.~\ref{fig:anchor}(b) shows how this anchor is established for different anatomical targets: the initial probe pose is defined at the inferior xiphoid tip for liver and gallbladder scanning, at the T12 surface landmark for left-kidney scanning, and at the wrist area for the radial artery scanning. Although the subsequent scanning trajectories differ across tasks, each is represented relative to its anatomically defined starting pose.


\subsubsection{Multimodal Observation Encoding}
\label{sec:multimodal_observation_encoding}

At each policy update $t$, the model receives a synchronized history of $K$ observations,

\begin{equation}
\mathcal{O}_{t}
=
\left\{
\mathbf{I}^{\mathrm{US}}_{t-K+1:t},
\mathbf{I}^{\mathrm{W}}_{t-K+1:t},
\mathbf{x}^{A}_{t-K+1:t},
\mathbf{f}^{z}_{t-K+1:t}
\right\},
\label{eq:multimodal_observation_history}
\end{equation}

where $\mathbf{I}^{\mathrm{US}}$ denotes the ultrasound image, $\mathbf{I}^{\mathrm{W}}$ the image acquired by the wrist camera, $\mathbf{x}^{A}$ the anchor-relative probe state as defined in Eq.~\eqref{eq:anchor_relative_state}, and $\mathbf{f}^{z}$ the normal component of the probe-tissue contact force. During both training and robotic execution, the force measurements are synchronized with the corresponding ultrasound frames.

The four modalities provide complementary information for ultrasound navigation. The ultrasound images directly visualize the internal anatomy and provide the primary feedback for approaching the target view. The wrist-camera images capture the local relationship between the probe and the patient surface. The anchor-relative probe states describe the spatial progression of the scan with respect to its initial anatomical reference. Finally, the contact force characterizes the current probe-tissue interaction and helps the policy account for changes in surface deformation and acoustic contact.

\begin{figure*}
    \centering
    \includegraphics[width=0.9\linewidth]{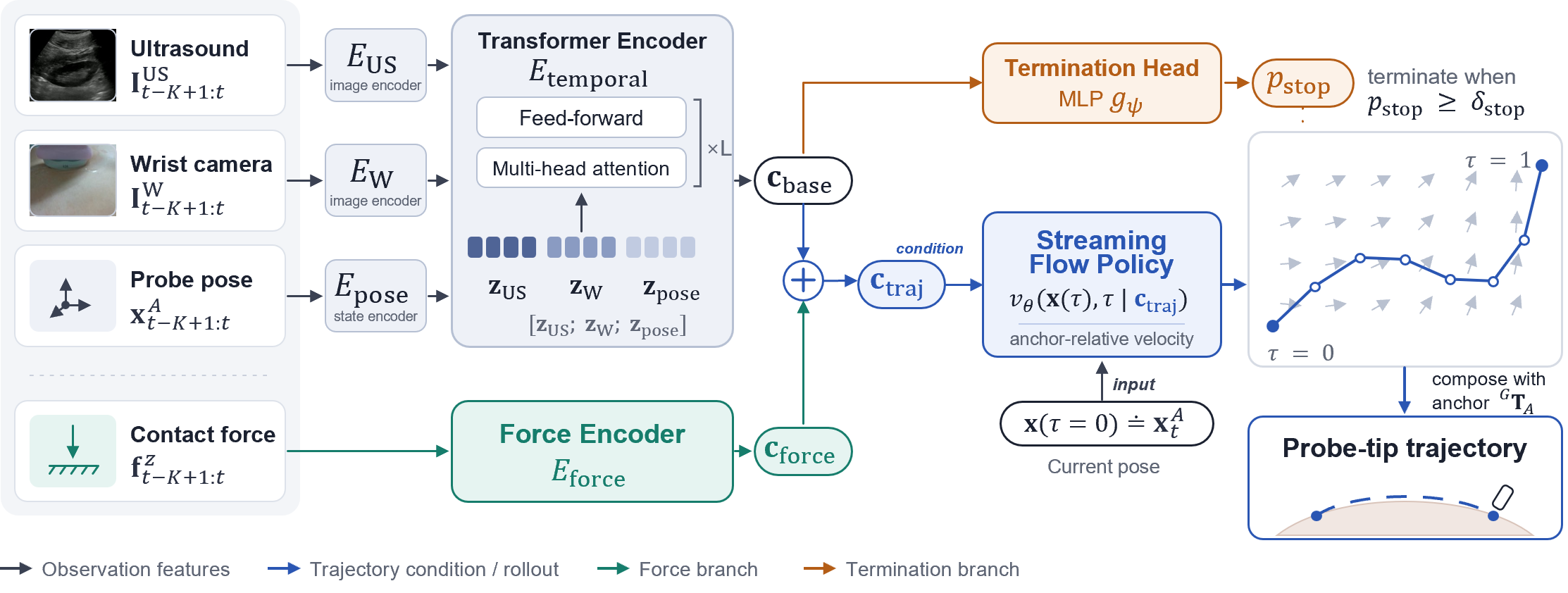}
    \caption{Architecture of the anchor-relative Streaming Flow Policy. Ultrasound images, wrist-camera images, and anchor-relative probe states are fused by a shared Transformer encoder into $\mathbf{c}_{\mathrm{base}}$, and a separate force branch adds the residual $\mathbf{c}_{\mathrm{force}}$ to form the trajectory condition $\mathbf{c}_{\mathrm{traj}}$. Conditioned on $\mathbf{c}_{\mathrm{traj}}$, the velocity field $v_\theta$ is integrated from the current pose $\mathbf{x}^{A}_{t}$ over $\tau \in [0, 1]$ and composed with the episode anchor ${}^{G}\mathbf{T}_{A}$ to produce the probe-tip trajectory. A termination head $g_\psi$ predicts $p_{\mathrm{stop}}$ from the stop-gradient copy $\mathrm{sg}(\mathbf{c}_{\mathrm{base}})$.}
    \label{fig:network}
\end{figure*}

As shown in Fig.~\ref{fig:network}, the ultrasound images, wrist-camera images, and anchor-relative probe states are first processed by modality-specific encoders $E_{\mathrm{US}}$, $E_{\mathrm{W}}$, and $E_{\mathrm{pose}}$, yielding features $\mathbf{z}_{\mathrm{US}}$, $\mathbf{z}_{\mathrm{W}}$, and $\mathbf{z}_{\mathrm{pose}}$. These features are then fused and aggregated by a shared Transformer encoder:

\begin{equation}
\mathbf{c}_{\mathrm{base}}
=
E_{\mathrm{temporal}}
\left(
[
\mathbf{z}_{\mathrm{US}};
\mathbf{z}_{\mathrm{W}};
\mathbf{z}_{\mathrm{pose}}
]
\right),
\label{eq:base_multimodal_condition}
\end{equation}

where $[\cdot;\cdot]$ denotes feature concatenation. The resulting representation jointly captures anatomical appearance, local surface context, and anchor-relative motion history.

The force history is processed through a separate force-encoding branch,
$\mathbf{c}_{\mathrm{force}}=E_{\mathrm{force}}\left(\mathbf{f}^{z}_{t-K+1:t}\right)$
The resulting force feature is incorporated as a residual into the base multimodal representation:
$\mathbf{c}_{\mathrm{traj}}=\mathbf{c}_{\mathrm{base}}+\mathbf{c}_{\mathrm{force}}$.
The complete condition $\mathbf{c}_{\mathrm{traj}}$ is used by the Streaming Flow Policy to predict the velocity field in the anchor-relative pose space. In contrast, the termination head uses the force-independent representation $\mathbf{c}_{\mathrm{base}}$, as described in Sec.~\ref{sec:termination_prediction}. This design allows contact force to influence probe-motion generation without making the termination decision directly dependent on instantaneous force variations.


\subsubsection{Anchor-Relative Streaming Flow Learning}
\label{sec:anchor_relative_flow_learning}

We employ Streaming Flow Policy (SFP)
\cite{jiang2025streaming} to model the evolution of the probe state in the anchor-relative pose space. Before being processed by the flow model, the anchor-relative pose vectors are normalized using statistics computed from the training set. For conciseness, the notation below refers to these normalized pose vectors.

Let $\tau\in[0,1]$ denote the local progress within an $H$-step prediction horizon, where $\tau=0$ and $\tau=1$ correspond to the beginning and end of the horizon, respectively. The variable $\tau$ does not represent the elapsed time of the complete scanning episode. Conditioned on $\mathbf{c}_{\mathrm{traj}}$, the network parameterizes a time-dependent velocity field $v_\theta$ that defines the evolution of the anchor-relative probe state:
\begin{equation}
\frac{\mathrm{d}\mathbf{x}(\tau)}{\mathrm{d}\tau}
=
v_{\theta}\left(\mathbf{x}(\tau),\tau \mid \mathbf{c}_{\mathrm{traj}}\right)
\in\mathbb{R}^{9}.
\label{eq:anchor_relative_flow}
\end{equation}


Following SFP, we construct a stabilizing target velocity field around each demonstrated trajectory. At each training iteration, an index $j\in\{0,\ldots,H\}$ is sampled uniformly from the demonstrated trajectory $\boldsymbol{\xi}_{t}$ as defined in Eq.~\eqref{eq:anchor_relative_trajectory}, with the associated trajectory progress $\tau_j=j/H$. A perturbed state is then sampled around the corresponding demonstrated state:

\begin{equation}
\widetilde{\mathbf{x}}_{j}
=
\boldsymbol{\xi}_{t,j}
+
\sigma_{0}\boldsymbol{\epsilon},
\qquad
\boldsymbol{\epsilon}
\sim
\mathcal{N}(\mathbf{0},\mathbf{I}),
\label{eq:perturbed_probe_state}
\end{equation}

where $\sigma_{0}$ controls the magnitude of the perturbation. Given the demonstrated trajectory derivative $\dot{\boldsymbol{\xi}}_{t,j}$, the corresponding target velocity is

\begin{equation}
\mathbf{v}^{*}_{j}
=
\dot{\boldsymbol{\xi}}_{t,j}
-
k
\left(
\widetilde{\mathbf{x}}_{j}
-
\boldsymbol{\xi}_{t,j}
\right),
\label{eq:stabilizing_target_velocity}
\end{equation}

where $k$ denotes the stabilization gain. The first term advances the state along the demonstrated trajectory, while the second term directs perturbed states back toward the demonstration. The latter allows the policy to learn corrective behavior around the demonstrated probe trajectories rather than only reproducing their nominal evolution.

The velocity network is trained by minimizing the discrepancy between the predicted and target velocities in Eq.~\eqref{eq:stabilizing_target_velocity}:

\begin{equation}
\mathcal{L}_{\mathrm{traj}}
=
\mathbb{E}_{j,\boldsymbol{\epsilon}}
\left[
\left\|
v_{\theta}
\left(
\widetilde{\mathbf{x}}_{j},
\tau_j
\mid
\mathbf{c}_{\mathrm{traj}}
\right)
-
\mathbf{v}^{*}_{j}
\right\|_{2}^{2}
\right].
\label{eq:trajectory_flow_loss}
\end{equation}

During inference, each flow rollout is initialized from the currently measured anchor-relative probe state at $\tau=0$. The velocity network is then evaluated sequentially at $\tau_i=i/H$, for $i=0,\ldots,H-1$, and its outputs are integrated to
generate $H$ future probe states. The multimodal condition $\mathbf{c}_{\mathrm{traj}}$ remains fixed throughout one rollout. Only the first $H_{exec}$ generated states are executed, where $H_{exec}$ denotes the execution-chunk length. The policy is subsequently queried again using updated observations and the latest measured probe state, and $\tau$ is reset to zero for the new rollout.


\subsubsection{Termination Prediction}
\label{sec:termination_prediction}

In addition to generating the probe trajectory, the policy predicts whether the target ultrasound view has been reached. This allows the duration of the navigation episode to adapt to the observed anatomical progress rather than being determined by a fixed trajectory length.

The termination branch uses the force-independent multimodal representation $\mathbf{c}_{\mathrm{base}}$ defined in Eq.~\eqref{eq:base_multimodal_condition}. In particular, the contact-force feature is excluded from this branch so that the stopping decision is based on the ultrasound appearance, wrist-camera context, and anchor-relative probe state rather than instantaneous variations in the contact force.
The termination head is implemented as a multilayer perceptron (MLP) that takes the detached force-independent representation as input and outputs the probability that the target view has been reached:

\begin{equation}
p_{\mathrm{stop}}
=
g_{\psi}
\left(
\operatorname{sg}
\left(
\mathbf{c}_{\mathrm{base}}
\right)
\right),
\label{eq:stop_probability}
\end{equation}

where $\operatorname{sg}(\cdot)$ denotes the stop-gradient operation and $g_{\psi}$ denotes the MLP-based termination head.

Let $y_{\mathrm{stop}}\in\{0,1\}$ denote the termination label, with $y_{\mathrm{stop}}=1$ indicating that the target view has been reached. Because termination samples occur less frequently than non-termination samples, we employ a positively weighted binary cross-entropy loss:

\begin{equation}
\mathcal{L}_{\mathrm{stop}}
=
-
w_{\mathrm{pos}}y_{\mathrm{stop}}
\log p_{\mathrm{stop}}
-
\left(
1-y_{\mathrm{stop}}
\right)
\log
\left(
1-p_{\mathrm{stop}}
\right),
\label{eq:stop_loss}
\end{equation}

where $w_{\mathrm{pos}}$ controls the contribution of positive termination samples. The complete training objective combines Eq.~\eqref{eq:trajectory_flow_loss} and Eq.~\eqref{eq:stop_loss}: $\mathcal{L}=\mathcal{L}_{\mathrm{traj}}+\lambda_{\mathrm{stop}}\mathcal{L}_{\mathrm{stop}}$, where $\lambda_{\mathrm{stop}}$ balances trajectory learning and termination prediction.

Although the trajectory and termination objectives are optimized jointly, applying the stop-gradient operation to $\mathbf{c}_{\mathrm{base}}$ prevents gradients of $\mathcal{L}_{\mathrm{stop}}$ from propagating into the shared multimodal encoder. Consequently, the termination loss updates only the MLP-based termination head $g_{\psi}$, whereas the trajectory loss updates the shared multimodal encoder, the force-encoding branch, and the velocity-field network.


\subsubsection{Receding-Horizon Policy Execution}
\label{sec:receding_horizon_execution}

During robotic execution, the anchor frame remains fixed at the probe-tip pose established at the beginning of the navigation episode. At each policy update, the latest multimodal observation history is encoded once to obtain the trajectory condition $\mathbf{c}_{\mathrm{traj}}$. Starting from the currently measured anchor-relative probe state $\mathbf{x}^{A}_{t}$, the velocity network is evaluated sequentially over the $H$-step prediction horizon. Using Euler integration of Eq.~\eqref{eq:anchor_relative_flow}, the predicted state at each step is updated as
\begin{equation}
\widehat{\mathbf{x}}^{A}_{t+i+1}
=
\widehat{\mathbf{x}}^{A}_{t+i}
+
\Delta\tau\,
v_{\theta}
\left(
\widehat{\mathbf{x}}^{A}_{t+i},
\tau_i
\mid
\mathbf{c}_{\mathrm{traj}}
\right),
\label{eq:flow_euler_integration}
\end{equation}
where $\widehat{\mathbf{x}}^{A}_{t}=\mathbf{x}^{A}_{t}$, $\tau_i=i/H$, and $\Delta\tau=1/H$ for $i=0,\ldots,H-1$. The trajectory condition remains fixed throughout one rollout, while the predicted probe state and horizon progress are updated at each integration step. Consequently, the observation encoder is evaluated once per rollout, whereas the velocity network is evaluated $H$ times.

For robotic execution, each predicted anchor-relative pose is composed with the fixed anchor pose to recover the corresponding absolute probe-tip target:
\begin{equation}
{}^{G}\widehat{\mathbf{T}}_{P_{t+i}}
=
{}^{G}\mathbf{T}_{A}
{}^{A}\widehat{\mathbf{T}}_{P_{t+i}},
\label{eq:relative_to_absolute_pose}
\end{equation}
where $\mathcal{F}_{G}$ corresponds to the robot-base frame during robotic roll-outs.

Although the policy generates $H$ future states, only the first $H_{exec}$ states are executed, where $H_{exec}\leq H$ is the execution-chunk length. These target poses are issued sequentially to the robot controller. Once the execution chunk has been completed, the policy is queried again using the updated multimodal observations and the latest measured probe state. The local horizon progress $\tau$ is then reset to zero for the new rollout. This receding-horizon procedure continuously incorporates new ultrasound, visual, pose, and force feedback while preventing the robot from executing the complete predicted trajectory open-loop.

The termination probability $p_{\mathrm{stop}}$ in Eq.~\eqref{eq:stop_probability} is evaluated at each policy update. A stop prediction is considered positive when $p_{\mathrm{stop}} \geq \delta_{\mathrm{stop}}$, where $\delta_{\mathrm{stop}}$ is a predefined probability threshold. To reduce sensitivity to isolated predictions, termination is confirmed only when the threshold is exceeded for a predefined number of consecutive policy updates. Once termination is confirmed, the remaining target poses are discarded, the stopping state is latched, and the robot holds the currently measured probe pose.

Before being issued to the robot controller, the generated target poses are additionally constrained by predefined translational and rotational step limits and smoothed relative to the previously commanded pose. These safeguards prevent abrupt probe motions caused by isolated prediction errors during real-time execution.

\subsubsection{Post-Scan Best-View Selection}
\label{sec:best_view_selection}

For abdominal scanning tasks, we additionally perform automatic best-view selection after the scan has terminated. While the termination head determines when to stop robotic scanning, this module retrospectively identifies the preferred view from the recorded ultrasound sequence. Post-scan selection allows candidate views acquired throughout the scan to be compared, rather than restricting the output to the image observed at termination. The module is trained independently of the scanning policy and operates exclusively on ultrasound images.

Each frame is resized to $256\times256$ pixels and its intensity values are scaled to $[0,1]$. A frozen pretrained EchoCare~\cite{zhang2025fully} Swin Transformer extracts hierarchical image features, which are spatially pooled and concatenated before being passed to an organ-specific residual MLP. Separate heads are trained for the kidney, gallbladder, and liver, each producing a scalar frame-ranking score. Training uses manually selected best-view frames as supervision and combines a listwise cross-entropy loss with a pairwise ranking loss against high-scoring competing frames from the same video. Temporally adjacent frames are excluded from these comparisons to reduce penalization of visually similar views.

For an ultrasound frame $I_t$, the organ-specific ranking score is computed as
\begin{equation}
s_t = h_o\!\left(f_{\mathrm{EC}}(I_t)\right),
\qquad
t^{\star} = \operatorname*{arg\,max}_{t\in\mathcal{C}} s_t,
\label{eq:best_view_selection}
\end{equation}
where $f_{\mathrm{EC}}$ denotes the frozen EchoCare~\cite{zhang2025fully} encoder followed by feature pooling and concatenation, $h_o$ is the ranking head for organ $o$, and $\mathcal{C}$ contains the frame indices evaluated during the coarse-to-fine search. The selected frame is $I_{t^{\star}}$.

At inference, a coarse-to-fine search is performed over the recorded video: every eighth frame is initially scored, followed by frame-by-frame refinement within eight frames of each of the three highest-scoring candidates. The highest-ranked frame among all evaluated candidates is exported at its original resolution for subsequent physician assessment. Frame selection requires no manual annotations at inference, and the ranking score expresses relative preference within the video rather than calibrated clinical image quality.

\begin{figure}
    \centering
    \includegraphics[width=0.8\linewidth]{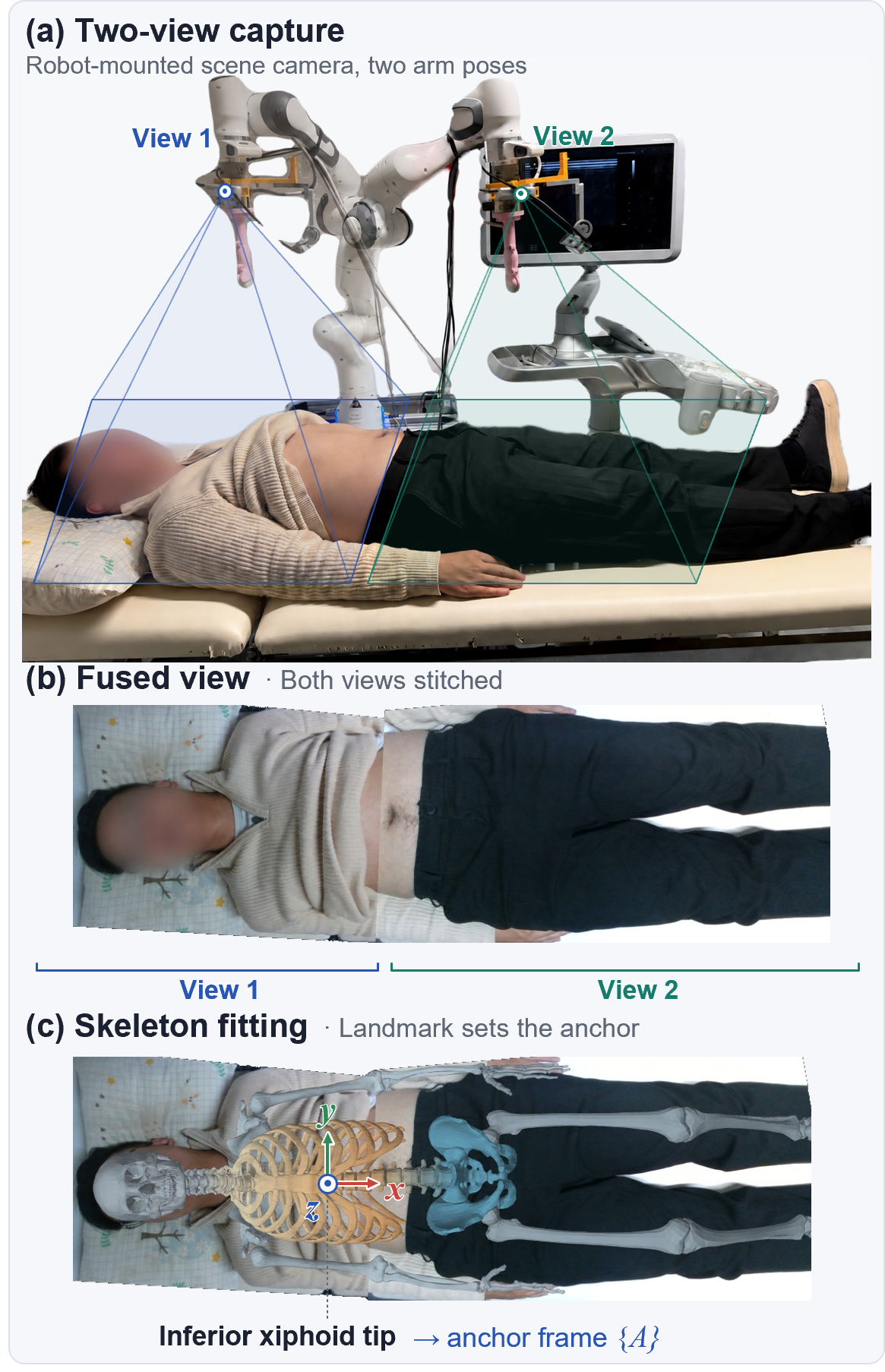}
    \caption{Anatomical landmark localization for automatic probe placement. (a) The scene camera captures two overlapping views of the participant. (b) The views are stitched into a whole-body image. (c) HSMR reconstructs the underlying skeleton to identify the target landmark, illustrated here by the inferior xiphoid tip, which guides initial probe placement and subsequent anatomical anchor definition.}
    \label{fig:hsmr}
\end{figure}

\subsection{Anatomy-Guided Placement Module}
\label{sec:placement_module}

Before ultrasound navigation, a rule-based placement module positions the probe at a task-specific anatomical starting point. The module identifies the relevant anatomical feature from an RGB-D observation of the participant, maps it onto the measured body surface, and determines an initial probe pose. This pose establishes contact with the patient and defines the anatomical anchor frame for the learned navigation policy. No imitation learning is involved in this stage.

\subsubsection{Anatomical Target Localization}
For the abdominal tasks, the robot moves the scene camera to predefined observation poses from which the subject's body can be completely observed. When the body cannot be covered by a single image, multiple overlapping RGB images are acquired and combined into a unified whole-body view, as shown in Fig.~\ref{fig:hsmr}(a) and (b). A depth frame is also recorded together with the camera intrinsics and the camera-to-robot transformation.

Given the whole-body RGB image, HSMR~\cite{xia2025reconstructing} reconstructs a biomechanically constrained human model comprising the external body shape and its underlying skeletal structure, as shown in Fig.~\ref{fig:hsmr}(c). For each abdominal task, a predefined point-based anatomical feature is extracted from the reconstructed skeleton. The target may correspond to the image projection of a skeletal landmark or to a point defined relative to multiple landmarks, allowing the placement location to adapt to individual body proportions.

For radial artery sweeping, the target lies on the forearm, where whole-body skeletal reconstruction is unnecessary. Instead, an RGB-D image of the forearm is acquired from a predefined observation pose, and MediaPipe Hands~\cite{zhang2020mediapipe} is applied to the RGB image to detect the hand landmarks. The detected wrist landmark is used as the target point for probe placement.

In both cases, the extracted anatomical point is mapped into the aligned depth image and back-projected into 3-D. It is then transformed into the robot base frame and associated with the nearest valid point in the measured surface point cloud. In this way, HSMR or MediaPipe Hands identifies the anatomical location relevant to the task, while the RGB-D observation determines its metric position in the robot workspace.

\subsubsection{Placement Pose Generation}

The local surface normal is estimated by principal component analysis of a neighborhood $\mathcal{N}_k$ around the target surface point, as the eigenvector associated with the smallest eigenvalue of its covariance matrix, with its sign selected to point outward from the body~\cite{bi2026scanning}.

The probe axis is aligned with the surface normal to approach the skin approximately perpendicularly, while the rotation about this axis is assigned according to the scanning task. The target point and probe orientation jointly define the initial placement pose. After contact is established, the resulting probe-tip pose is fixed as the anatomical anchor frame, and control is transferred to the learned ultrasound-navigation policy.


\section{Experiments}
We evaluate the proposed framework through a series of experiments with human participants. After describing the hardware setup, data acquisition, and evaluation protocol (Sec.~\ref{subsec:hardware}-\ref{subsec:evaluation_protocol}), we present four groups of experiments. First, we compare the learned scanning policy with a fine-tuned $\pi_0$ model on previously unseen participants (Sec.~\ref{subsec:baseline}). Second, we conduct ablation studies on the wrist-camera input, force input, and pose representation (Sec.~\ref{subsec:ablation}). Third, we validate the complete pipeline, combining automatic anatomy-guided probe placement with learned navigation (Sec.~\ref{subsec:full_system}). Finally, we deploy the same policies on two different robots, a Franka Emika Panda and a KUKA LBR iiwa 7 R800, to assess cross-embodiment transfer (Sec.~\ref{subsec:cross_embodiment}).

\subsection{Hardware Setup}
\label{subsec:hardware}

\begin{figure}
    \centering
    \includegraphics[width=\linewidth]{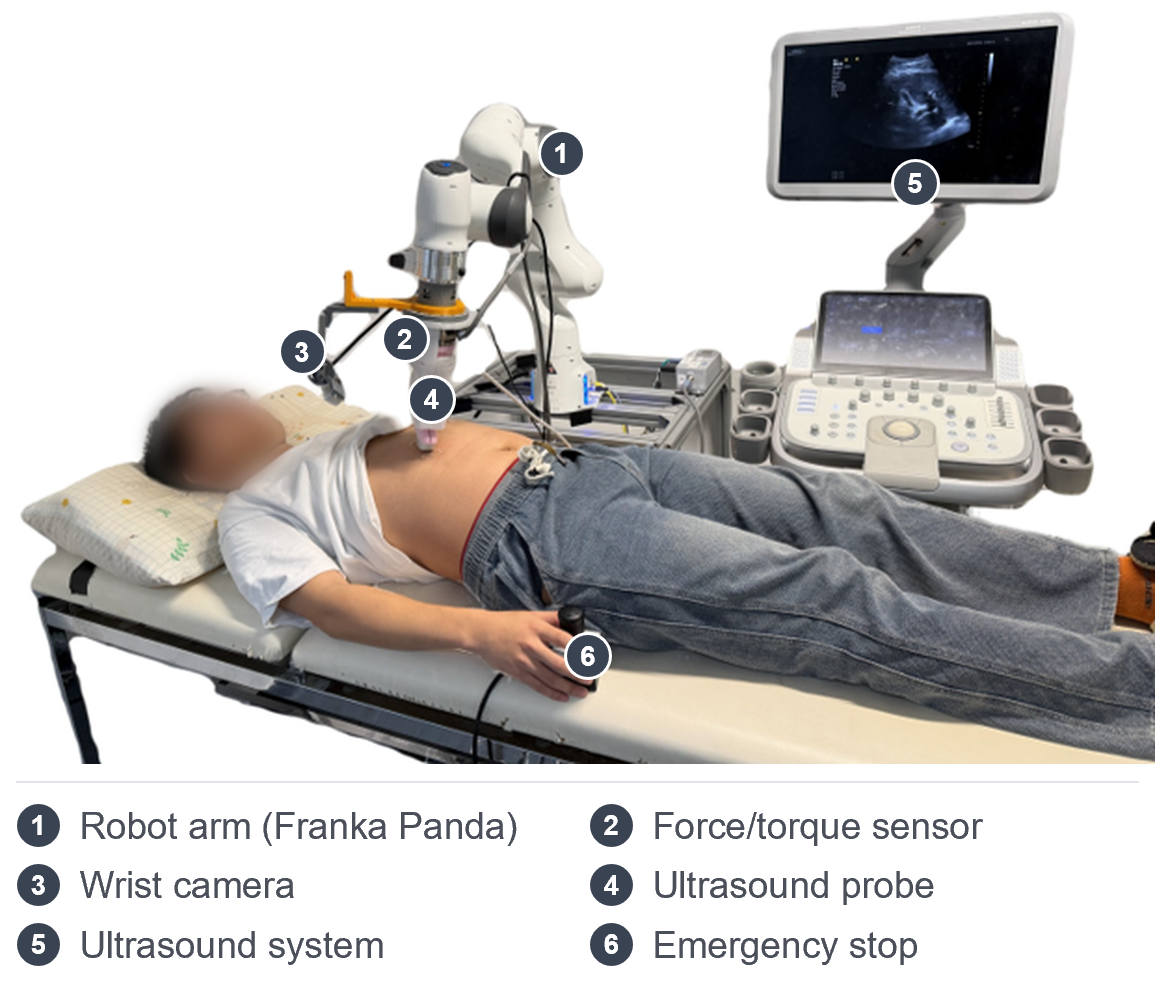}
    \caption{Experimental setup for autonomous robotic ultrasound scanning. A 7-DoF robot manipulates the ultrasound probe, with a wrist camera and force/torque sensor providing visual and contact feedback. The participant holds an emergency-stop button throughout the procedure.}
    \label{fig:image_scene}
\end{figure}

The hardware setup for the experiment is shown in Fig.~\ref{fig:image_scene}. A Siemens ACUSON Juniper ultrasound system is used with a 5C1 convex probe for liver, gallbladder, and kidney scanning and a 12L3 linear probe for radial artery scanning. The B-mode ultrasound images are captured from the ultrasound machine via a frame grabber (MAGEWELL) at a frequency of 30 Hz. A six-axis force/torque sensor (GAMMA, SI-32-2.5/SI-65-5, ATI) is mounted at the end of the ultrasound probe holder to measure contact forces during demonstration recording and provide force feedback for control during robotic rollouts. Two cameras (Intel RealSense D435) are used: one observes the probe footprint to capture the interaction between the probe and the tissue surface, while the other captures a whole-body view and the scanning region's 3D surface geometry to guide initial probe placement. During free-hand demonstration recording, an optical tracking system (Polaris Vicra, NDI) records the movements of the probe. The acquisition platform can be mounted directly onto the end effector of a seven-degree-of-freedom robotic manipulator (Franka Emika Panda, Franka GmbH or KUKA LBR iiwa 7 R800, KUKA, Germany) to execute the learned scanning policies. Robot control and data acquisition are implemented using ROS. A workstation with Ubuntu 20.04 system equipped with an NVIDIA RTX PRO 6000 Blackwell GPU handles system control and all model training. All the models are trained using the AdamW optimizer with a learning rate of $1\times10^{-4}$ (cosine schedule with a 500-step warm-up) and a batch size of 64 for 800 epochs.

\textit{Implementation details:} The policy uses an observation history of $K=3$ and a prediction horizon of $H=16$, of which the first $H_{\mathrm{exec}}=8$ predicted states are executed before the next policy update. During training, perturbations are sampled with $\sigma_0=0.01$ in the normalized pose space, and the stabilization gain is set to $k=1.0$. The termination loss uses a positive weight of $w_{\mathrm{pos}}=5.0$ and is weighted by $\lambda_{\mathrm{stop}}=0.5$. At inference, termination is confirmed when $p_{\mathrm{stop}}\geq\delta_{\mathrm{stop}}=0.5$. Target poses are sent to the Cartesian impedance controller at 30~Hz.

\begin{table*}[t]
\centering
\caption{Demonstration collection across scanning tasks.
For abdominal tasks, the long axis of the probe footprint is
initially aligned with the craniocaudal direction; the forearm
task is a cross-sectional sweep.
Volunteer counts are reported per task.}
\label{tab:demonstration_tasks}
\small
\renewcommand{\arraystretch}{1.15}
\begin{tabular}{@{} l l r r l @{\hspace{0.8em}$\rightarrow$\hspace{0.8em}} l @{}}
\toprule
Region & Task & Demos. & Volunteers & Starting condition & Ending condition \\
\midrule
\multirow{3}{*}{Abdomen}
  & Kidney        & 69 & 5 & T12 surface landmark & Left kidney, longitudinal view \\
  & Gallbladder   & 60 & 5 & Inferior xiphoid tip & Gallbladder, longitudinal view \\
  & Liver         & 82 & 5 & Inferior xiphoid tip & Left portal vein with segmental branches \\
\midrule
Forearm
  & Radial artery & 83 & 2 & Wrist                & Bifurcation of radial and ulnar arteries \\
\bottomrule
\end{tabular}
\end{table*}

\subsection{Data Acquisition}
\label{subsec:data_acquisition}
We test our proposed system on four different ultrasound scanning tasks on healthy volunteers, including the standard plane navigation task for liver, gallbladder, and kidney, and the long-sweep task for radial artery scanning. 
Human-participant data acquisition and real-robot evaluation were conducted under institutional ethics approval.\footnote{Ethics Committee of the University, reference no.~xxx/xx.}

We collected 294 demonstrations across four scanning tasks, comprising three standard-plane acquisition tasks and one long-horizon sweep task. The standard-plane tasks target a longitudinal view of the left kidney, a longitudinal view of the gallbladder, and a view of the left portal vein with its segmental branches, supporting the assessment of renal morphology and abnormalities, gallbladder morphology and abnormalities, and left-lobe anatomy and lesion localization, respectively. In contrast, the radial artery task is a continuous sweep that follows the artery in cross-section from the wrist to the elbow, where the brachial artery bifurcates into the radial and ulnar arteries.
Each demonstration begins at a predefined anatomical landmark and ends upon reaching the target plane or completing the prescribed sweep. For the liver, gallbladder, and kidney tasks, the long axis of the probe footprint is initially aligned with the craniocaudal direction. For liver and gallbladder demonstrations, volunteers are instructed to take a deep breath and hold it during recording. Tab.~\ref{tab:demonstration_tasks} summarizes the demonstration counts, volunteer numbers, and task-specific starting and ending conditions.

All demonstrations are collected using the proposed ultrasound data acquisition platform, which records ultrasound images, wrist-camera images, force measurements, and tracked probe poses. Recordings with tracking failures are excluded. The remaining recordings are temporally trimmed based on the ultrasound stream to retain valid imaging intervals, removing the initial and final periods during which the probe establishes or loses contact with the tissue surface. Each demonstration is also manually annotated to identify the earliest frame at which scanning can terminate, corresponding to successful acquisition of the target standard plane or completion of the prescribed sweep.

\subsection{Evaluation Protocol}
\label{subsec:evaluation_protocol}
We evaluate performance using six metrics:

(1) \textbf{task success rate}, defined as the proportion of all attempted rollouts that complete the prescribed vascular scan for the arm or acquire a qualifying target view for abdominal organs. For radial artery scanning, success requires autonomous tracking of the target vessel over at least 95\% of the prescribed scan extent. Task success is evaluated independently of autonomous stopping;

(2) \textbf{ROI image quality}. For the abdominal tasks, the quality of the acquired target view is evaluated by physicians using a five-point rating scale. The assessment considers (i) the visibility of task-relevant anatomical structures, (ii) the adequacy of the intended imaging plane, (iii) structural clarity and the presence of imaging artifacts, and (iv) overall clinical interpretability. A score of 1 indicates an unusable image in which the target anatomy cannot be reliably identified, whereas a score of 5 indicates an excellent target view with clearly visualized anatomical landmarks, appropriate imaging geometry, minimal artifacts, and high clinical interpretability. The evaluation is performed independently by six physicians, including five senior physicians and one junior physician. 
For both methods, the rated frame of each rollout is selected from the recorded sequence by the same post-scan best-view selection module (Sec.~\ref{sec:best_view_selection}), and the physicians are blinded to the method that produced each sample.
Each physician evaluates the same set of 180 abdominal target-view samples, and the ratings are aggregated across raters for subsequent analysis. For radial artery scanning, where image quality must be maintained continuously along the scanning trajectory rather than at a single target view, scan quality is quantified by the \textit{vessel visibility ratio}, defined as the proportion of evaluated frames in which the target vessel lumen and its boundaries are clearly identifiable.

(3) \textbf{autonomous stopping success rate}, defined as the proportion of task-successful rollouts in which the robot stops appropriately after task completion without manual intervention;

(4) \textbf{time-to-target}, measured from rollout initiation to the first qualifying target view for abdominal tasks or completion of the prescribed scan for the arm;

(5) \textbf{mean peak normal contact force}, computed by first taking the maximum magnitude of the probe-normal contact force within each rollout and then averaging these peak values across all successful rollouts; 

(6) \textbf{translational jerk RMS}, computed for each rollout as the root-mean-square(RMS) Euclidean norm of the third time derivative of the measured probe-tip position.

Task success rate is computed over all attempted rollouts. Time-to-target, peak normal contact force, and autonomous stopping performance are evaluated only for successful rollouts to characterize execution performance conditional on task completion. This restriction accounts for differences in scan duration and coverage arising from incomplete or prematurely terminated scans. The three continuous metrics are computed separately for each successful rollout and reported as mean $\pm$ standard deviation. Autonomous stopping success rate is calculated as the proportion of successful rollouts in which the robot stops appropriately after task completion without manual intervention; autonomous stopping is not required to be identified as successful.

\begin{table*}[t]
\centering
\caption{Comparison of the proposed pipeline with fine-tuned $\pi_0$ across scanning tasks on unseen participants.}
\label{tab:baseline_comparison}
\renewcommand{\arraystretch}{1.2}
\setlength{\tabcolsep}{2pt}
\footnotesize

\begin{tabular*}{\textwidth}{@{\extracolsep{\fill}}lllccccccc@{}}
\toprule
\textbf{Region} &
\textbf{Task} &
\textbf{Method} &
\textbf{Rollouts} &
\shortstack{\textbf{Task success}\\\textbf{(\%) $\uparrow$}} &
\shortstack{\textbf{ROI quality}\\\textbf{ $\uparrow$}} &
\shortstack{\textbf{Autonomous}\\\textbf{stop (\%) $\uparrow$}} &
\shortstack{\textbf{Time-to-target}\\\textbf{(s) $\downarrow$}} &
\shortstack{\textbf{Peak force}\\\textbf{(N)}} &
\shortstack{\textbf{Jerk}\\$\mathrm{(m/s^3)} \downarrow$} \\
\midrule

\multirow{6}{*}{Abdomen}
& \multirow{2}{*}{Kidney}
& Fine-tuned $\pi_0$
& 50 & 86.0 & $2.78 \pm 1.22$ & 23.3
& $11.2 \pm 4.0$
& $7.2 \pm 8.3$
& $0.444 \pm 0.228$ \\
& & Anatomy-anchored SFP
& 50 & 96.0 & $3.47 \pm 1.09$ & 97.9
& $5.7 \pm 2.7$
& $4.3 \pm 2.0$
& $0.292 \pm 0.078$ \\

\cmidrule(l){2-10}

& \multirow{2}{*}{Gallbladder}
& Fine-tuned $\pi_0$
& 50 & 52.0 & $2.19 \pm 1.28$ & 19.2
& $18.1 \pm 8.7$
& $4.4 \pm 2.1$
& $0.376 \pm 0.067$ \\
& & Anatomy-anchored SFP
& 50 & 94.0 & $3.97 \pm 1.04$ & 93.6
& $9.7 \pm 5.5$
& $3.3 \pm 1.8$
& $0.262 \pm 0.047$ \\

\cmidrule(l){2-10}

& \multirow{2}{*}{Liver}
& Fine-tuned $\pi_0$
& 50 & 58.0 & $2.64 \pm 1.39$ & 17.2
& $12.8 \pm 7.7$
& $4.7 \pm 2.2$
& $0.420 \pm 0.330$ \\
& & Anatomy-anchored SFP
& 50 & 88.0 & $3.78 \pm 1.00$ & 95.5
& $9.5 \pm 5.3$
& $3.8 \pm 2.0$
& $0.217 \pm 0.031$ \\

\midrule

\multirow{2}{*}{Forearm}
& \multirow{2}{*}{Radial artery}
& Fine-tuned $\pi_0$
& 50 & 70.0 & $65.9\%$ & 28.6
& $17.9 \pm 3.7$
& $2.9 \pm 0.6$
& $0.605 \pm 0.082$ \\
& & Anatomy-anchored SFP
& 50 & 94.0 & $99.6\%$ & 100.0
& $11.7 \pm 3.0$
& $4.4 \pm 1.2$
& $0.625 \pm 0.063$ \\

\bottomrule
\end{tabular*}
\end{table*}

\subsection{Navigation Policy Evaluation}
\label{subsec:baseline}

The quantitative experiments in this section evaluate the learned imitation learning scanning policy. In all trials, the probe is manually positioned at the initial scanning pose; the complete pipeline, including automatic placement, is evaluated separately in Sec.~\ref{subsec:full_system}. We evaluate on 16 previously unseen participants of diverse ethnic backgrounds and genders, with heights ranging from 160 to 190~cm. The evaluation covers four anatomical targets and two task types: target-view acquisition for abdominal organs (kidney, gallbladder, and liver) and vascular sweeping along the radial artery in the forearm. These tasks assess complementary capabilities, namely locating an organ-specific standard view and continuously following a vascular structure. As a baseline, we fine-tune the $\pi_0$ vision-language-action (VLA) model~\cite{black2024pi0} on the same demonstrations described in Sec.~\ref{subsec:data_acquisition}. Each method is evaluated over 50 rollouts per task.

Tab.~\ref{tab:baseline_comparison} compares the proposed Anatomy-anchored SFP with fine-tuned $\pi_0$. Our method achieves higher task success rates on all four tasks, reaching $96.0\%$, $94.0\%$, and $88.0\%$ for kidney, gallbladder, and liver scanning, respectively, and $94.0\%$ for radial artery sweeping. Kidney scanning yields the highest success rate for both methods, whereas gallbladder and liver scanning are more challenging, particularly for $\pi_0$. The largest improvement occurs in gallbladder scanning, where the success rate increases from $52.0\%$ to $94.0\%$, followed by liver scanning ($58.0\%$ to $88.0\%$) and radial artery sweeping ($70.0\%$ to $94.0\%$).

\begin{figure*}
    \centering
    \includegraphics[width=\linewidth]{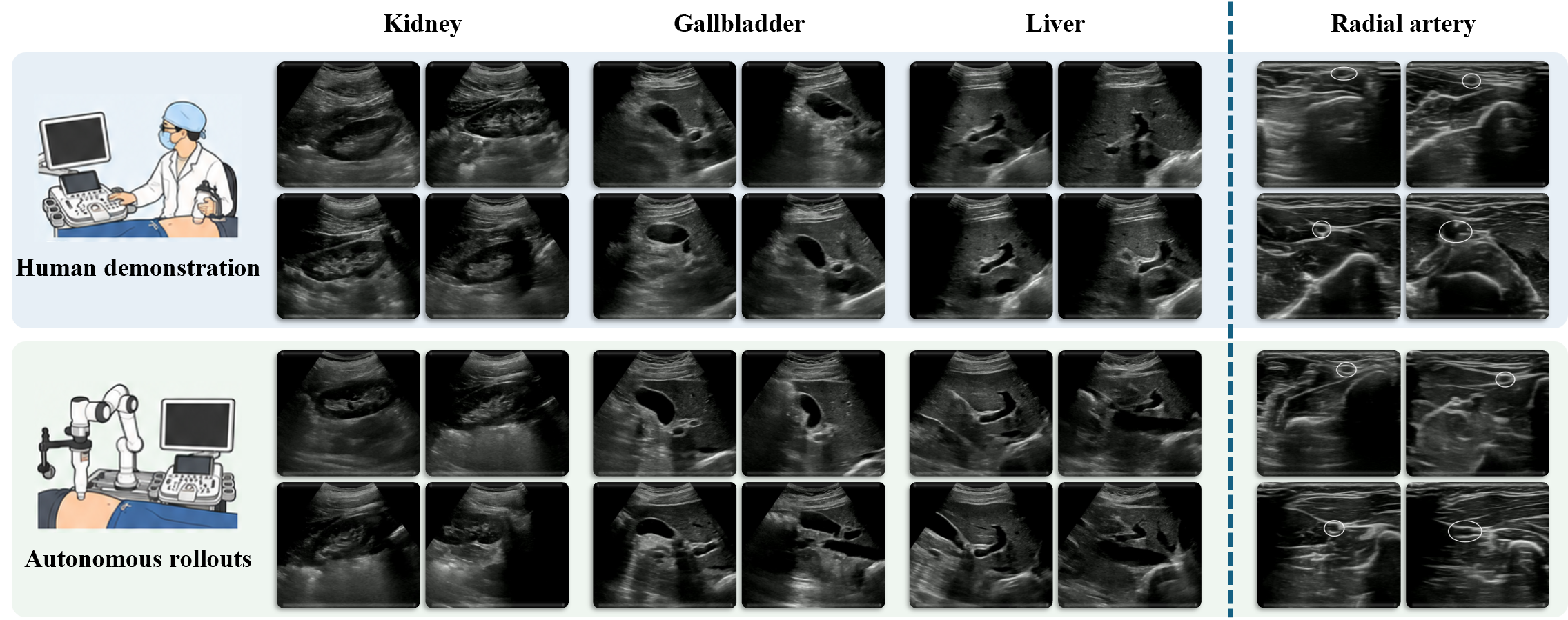}
    \caption{Representative ultrasound images from human demonstrations (top) and autonomous robotic rollouts (bottom). From left to right: kidney, gallbladder, and liver standard views, followed by images sampled along the radial artery sweep. White circles indicate the radial artery cross-section.}
    \label{fig:image_examples}
\end{figure*}

\begin{figure*}
    \centering
    \includegraphics[width=\linewidth]{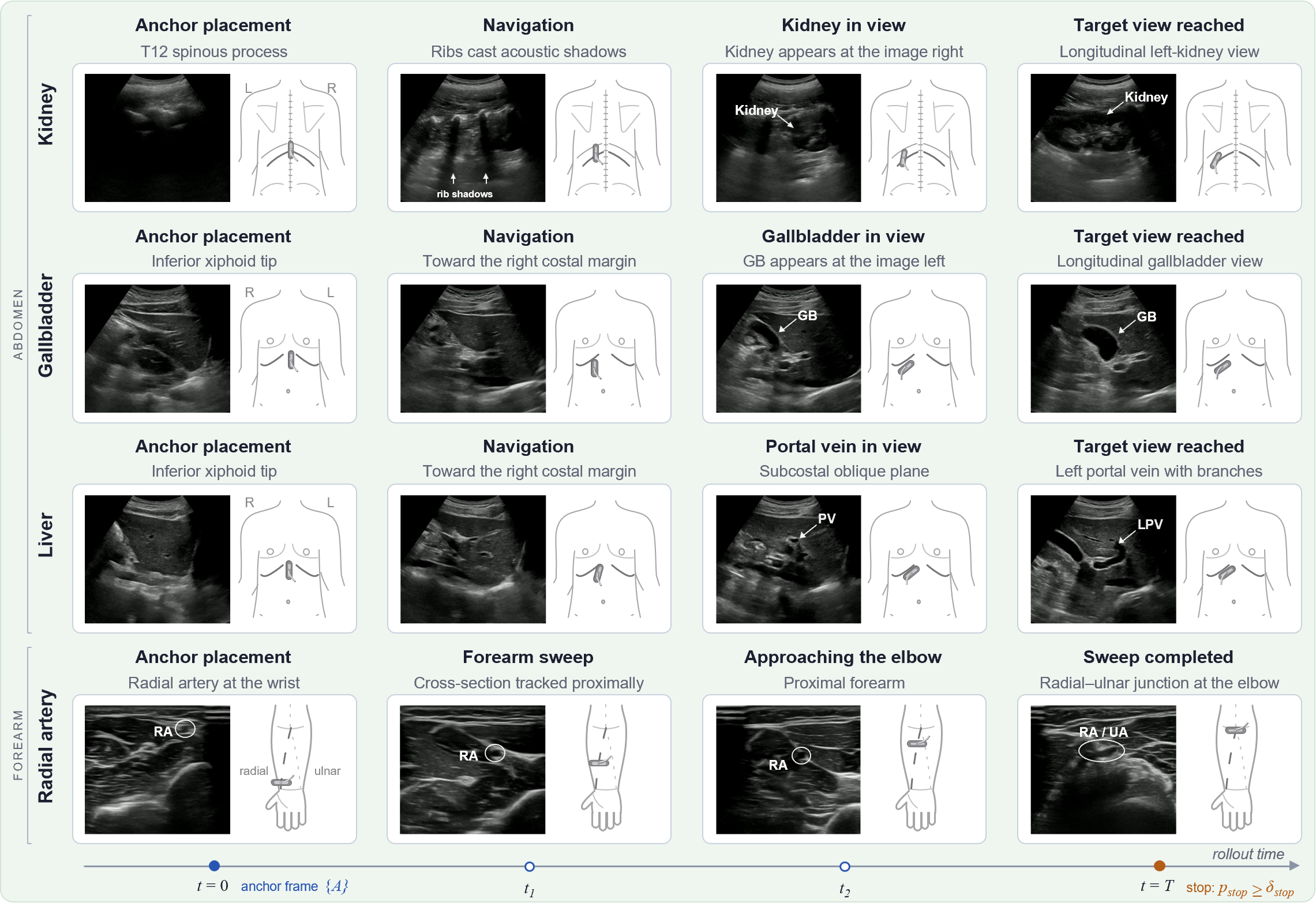}
    \caption{Representative autonomous rollouts of the Anatomy-anchored SFP
on unseen participants for the four scanning tasks: left-kidney, gallbladder, and liver
standard-plane acquisition (abdomen) and radial artery sweeping (forearm). Each row shows
four snapshots of one rollout. Each snapshot pairs the ultrasound image (left) with a
schematic of the corresponding probe pose on the body (right): anterior view for the
gallbladder and liver, posterior view for the kidney, and palmar view of the right forearm
for the radial artery. Each episode starts at the task-specific anatomical landmark, whose
probe pose defines the anchor frame $\{A\}$ ($t=0$), and ends when the termination head
predicts $p_\mathrm{stop} \ge \delta_\mathrm{stop}$ ($t=T$). The time axis is schematic, and
intermediate snapshots ($t_1$, $t_2$) are not equally spaced across tasks. Start and end
conditions follow Tab.~\ref{tab:demonstration_tasks}. GB: gallbladder; PV: portal vein; LPV: left
portal vein; RA: radial artery; UA: ulnar artery.}
    \label{fig:scanning_process}
\end{figure*}

Task success alone does not capture the diagnostic value of the acquired views. We therefore also examine the physician-rated ROI quality of the acquired target views. Our method achieves mean scores of $3.47$, $3.97$, and $3.78$ for kidney, gallbladder, and liver scanning, respectively, compared with $2.78$, $2.19$, and $2.64$ for $\pi_0$. Thus, the mean scores of our method exceed the scale midpoint of $3$ for all three organs, whereas those of $\pi_0$ remain below it. The quality improvement is largest for gallbladder scanning ($+1.78$), followed by liver ($+1.14$) and kidney ($+0.69$), mirroring the ordering of the success-rate gains. This indicates that the advantage of the proposed policy extends beyond reaching a qualifying view to the clarity and interpretability of the view itself. Notably, although $\pi_0$ reaches the kidney target in $86.0\%$ of rollouts, its kidney views are still rated lower, suggesting that reaching the vicinity of the target plane does not guarantee a well-oriented, artifact-free image. Kidney views also receive the lowest scores among the standard view tasks for our method, which may reflect the restricted acoustic window near the posterior ribs, where rib shadowing can partially obscure the renal outline (Fig.~\ref{fig:scanning_process}). In addition, our method yields smaller standard deviations across all three tasks ($1.00$-$1.09$ versus $1.22$-$1.39$), indicating more consistent view quality across participants and raters. For radial artery sweeping, the vessel visibility ratio increases from $65.9\%$ to $99.6\%$. Whereas $\pi_0$ loses clear delineation of the arterial lumen in roughly one-third of the evaluated frames, our policy keeps the vessel cross-section visible almost continuously throughout the sweep.

Beyond reaching the target, reliable task completion requires the policy to recognize when to stop. Although $\pi_0$ frequently reaches the target, particularly in kidney scanning, it stops appropriately without manual intervention in only $17.2$-$23.3\%$ of its successful abdominal rollouts and $28.6\%$ of its successful radial artery sweeps. In contrast, our method stops autonomously in $93.6$-$97.9\%$ of successful abdominal rollouts and in all successful radial artery sweeps. These results support the use of a dedicated termination head that decides completion from ultrasound appearance, wrist-camera context, and anchor-relative progress rather than leaving termination implicit in the action stream.

Our method also acquires targets more efficiently, reducing mean time-to-target by $26$-$49\%$ across the abdominal tasks and by $35\%$ for radial artery sweeping. On the abdominal tasks, these gains are accompanied by reductions of $19$-$40\%$ in mean peak contact force and $30$-$48\%$ in translational jerk, together with lower standard deviations in both metrics, indicating smoother and more consistent execution across trials. For radial artery sweeping, our method applies a higher mean peak force ($4.4$ vs.\ $2.9$~N) with comparable jerk ($0.625$ vs.\ $0.605$~m/s$^3$).

Fig.~\ref{fig:image_examples} shows representative ultrasound images from manual demonstrations and autonomous robotic rollouts. For the abdominal tasks, the rollout images are the frames chosen by the post-scan best-view selection module (Sec.~\ref{sec:best_view_selection}); for the radial artery task, images are sampled along the sweep. The robotic acquisitions capture the same task-relevant anatomical features as the demonstrations: the longitudinal profile of the left kidney, the longitudinal gallbladder lumen, the left portal vein with its segmental branches, and the radial artery cross-section (white circles), which remains in view as the probe advances along the forearm. Despite variations in anatomical appearance and acoustic conditions, these examples qualitatively illustrate the policy's ability to reach the task-relevant views through autonomous probe navigation, while the quality of the acquired views is quantified by the physician ratings and vessel visibility ratios in Tab.~\ref{tab:baseline_comparison}.

Fig.~\ref{fig:scanning_process} illustrates representative scanning sequences for the four tasks, pairing ultrasound observations with schematic probe poses. Starting from the task-specific anatomical anchor, the policy progressively adjusts the probe pose to acquire the required anatomical information. For kidney scanning, the sequence progresses from the T12 anchor through rib-shadowed views to the visualization of the kidney and the acquisition of its longitudinal view. Gallbladder and liver scanning both begin at the inferior xiphoid tip, followed by task-specific navigation toward the right costal margin to obtain the longitudinal gallbladder view and the left portal vein with its segmental branches, respectively. For radial artery scanning, the probe advances proximally from the wrist toward the bifurcation of radial and ulnar artery while following the arterial cross-section. These sequences illustrate both targeted standard view acquisition and extended anatomical sweeping within the proposed framework.

\begin{table*}[t]
\centering
\caption{Ablation study of wrist-camera input, force input, and pose representation on unseen participants.}
\label{tab:ablation}
\renewcommand{\arraystretch}{1.2}
\setlength{\tabcolsep}{2pt}
\footnotesize

\begin{tabular*}{\textwidth}{@{\extracolsep{\fill}}lllccccccc@{}}
\toprule
\textbf{Task} &
\textbf{Variant} &
\textbf{Configuration} &
\textbf{Rollouts} &
\shortstack{\textbf{Task success}\\\textbf{(\%) $\uparrow$}} &
\shortstack{\textbf{ROI quality}\\\textbf{$\uparrow$}} &
\shortstack{\textbf{Autonomous}\\\textbf{stop (\%) $\uparrow$}} &
\shortstack{\textbf{Time-to-target}\\\textbf{(s) $\downarrow$}} &
\shortstack{\textbf{Peak force}\\\textbf{(N)}} &
\shortstack{\textbf{Jerk}\\$\mathrm{(m/s^3)} \downarrow$} \\
\midrule

\multirow{6}{*}{Kidney}
& w/o wrist camera
& Nominal
& 30 & 76.7 & $2.79 \pm 1.17$ & 78.3
& $6.2 \pm 2.1$
& $5.3 \pm 2.5$
& $0.295 \pm 0.087$ \\

& w/o force input
& Nominal
& 30 & 86.7 & $3.07 \pm 1.08$ & 92.3
& $7.3 \pm 3.4$
& $5.8 \pm 5.5$
& $0.360 \pm 0.307$ \\

& Absolute pose
& Nominal
& 30 & 76.7 & $2.59 \pm 1.29$ & 73.9
& $7.0 \pm 2.5$
& $5.6 \pm 2.8$
& $0.372 \pm 0.153$ \\

& Absolute pose
& Rotated $45^\circ$
& 30 & 70.0 & $2.27 \pm 1.16$ & 71.4
& $6.9 \pm 2.7$
& $5.2 \pm 2.9$
& $0.443 \pm 0.189$ \\

& \textbf{Anatomy-anchored SFP}
& Nominal
& 50 & 96.0 & $3.47 \pm 1.09$ & 97.9
& $5.7 \pm 2.7$
& $4.3 \pm 2.0$
& $0.292 \pm 0.078$ \\

& \textbf{Anatomy-anchored SFP}
& Rotated $45^\circ$
& 30 & 93.3 & $3.54 \pm 1.15$ & 96.4
& $6.7 \pm 2.7$
& $4.8 \pm 2.0$
& $0.305 \pm 0.197$ \\

\midrule

\multirow{5}{*}{Radial artery}
& w/o wrist camera
& Rotated $90^\circ$
& 20 & 65.0 & $90.5\%$ & 69.2
& $12.8 \pm 2.9$
& $4.3 \pm 1.2$
& $0.590 \pm 0.093$ \\

& w/o force input
& Rotated $90^\circ$
& 20 & 90.0 & $97.6\%$ & 94.4
& $12.7 \pm 2.8$
& $5.4 \pm 1.0$
& $0.619 \pm 0.055$ \\

& Absolute pose
& Nominal
& 20 & 75.0 & $79.1\%$ & 73.3
& $20.0 \pm 3.5$
& $3.9 \pm 0.8$
& $0.572 \pm 0.072$ \\

& Absolute pose
& Rotated $90^\circ$
& 10 & 0.0 & -- & --
& --
& --
& -- \\

& \textbf{Anatomy-anchored SFP}
& Rotated $90^\circ$
& 50 & 94.0 & $99.6\%$ & 100.0
& $11.7 \pm 3.0$
& $4.4 \pm 1.2$
& $0.625 \pm 0.063$ \\

\bottomrule
\end{tabular*}
\end{table*}

\subsection{Ablation Studies}
\label{subsec:ablation}

We conduct ablation studies to assess the contribution of wrist-camera input, force input, and pose representation to scanning performance. Two representative tasks are selected, one from each scanning category: kidney standard-plane navigation and radial artery sweeping. Three variants of the proposed Anatomy-anchored SFP are considered: \textit{w/o wrist camera}, which removes wrist-camera observations; \textit{w/o force input}, which removes force input; and \textit{Absolute pose}, which replaces the anchor-relative pose representation with absolute poses in the robot-base frame. 
Because demonstrations are recorded in the optical tracker frame, the absolute robot-base trajectories used to train the \textit{Absolute pose} variant are obtained by composing each anchor-relative demonstration with a set of feasible end-effector poses selected around the task-specific starting region in the robot workspace. The variant therefore learns from the same demonstrated motions as the proposed model, expressed in absolute coordinates over a plausible range of starting poses.
All variants are retrained on the same demonstrations as the proposed model, with only the corresponding input modality removed or the pose representation changed.

To assess sensitivity to the scanning setup, rollouts are performed under two configurations. In the \textit{Nominal} configuration, the setup during robotic rollouts is aligned with that used during demonstration collection, whereas in the \textit{Rotated} configurations it is rotated by $45^\circ$ for kidney navigation and by $90^\circ$ for radial artery sweeping. The configuration of each evaluation is listed in Tab.~\ref{tab:ablation}, and each variant is compared with the proposed model under the same configuration wherever available.

Removing wrist-camera observations reduces task success from $96.0\%$ to $76.7\%$ for kidney navigation and from $94.0\%$ to $65.0\%$ for radial artery sweeping. Autonomous stopping among successful rollouts also decreases from $97.9\%$ to $78.3\%$ and from $100\%$ to $69.2\%$, respectively. The pronounced decline in stopping for radial artery sweeping is consistent with the loss of a direct visual cue: the wrist camera observes the shape of the arm, helping the policy recognize the endpoint of the sweep. For kidney navigation, the local surface view likely provides complementary information about the probe's location relative to the surrounding anatomy, supporting spatial reasoning beyond ultrasound images and anchor-relative poses alone. This is also reflected in view quality: the mean ROI score decreases from $3.47$ to $2.79$, so without the wrist camera the policy not only reaches qualifying views less often, but the views it acquires are also rated lower. For radial artery sweeping, the vessel visibility ratio decreases more moderately, from $99.6\%$ to $90.5\%$. Relative to the drop in task success, this modest decrease suggests that ultrasound feedback remains the primary cue for keeping the vessel in view, while the wrist camera mainly helps the policy follow sweep progress and recognize its endpoint. Overall, these results indicate that wrist-camera observations contribute both to local navigation and to recognition of task completion.

\begin{figure}
    \centering
    \includegraphics[width=\linewidth]{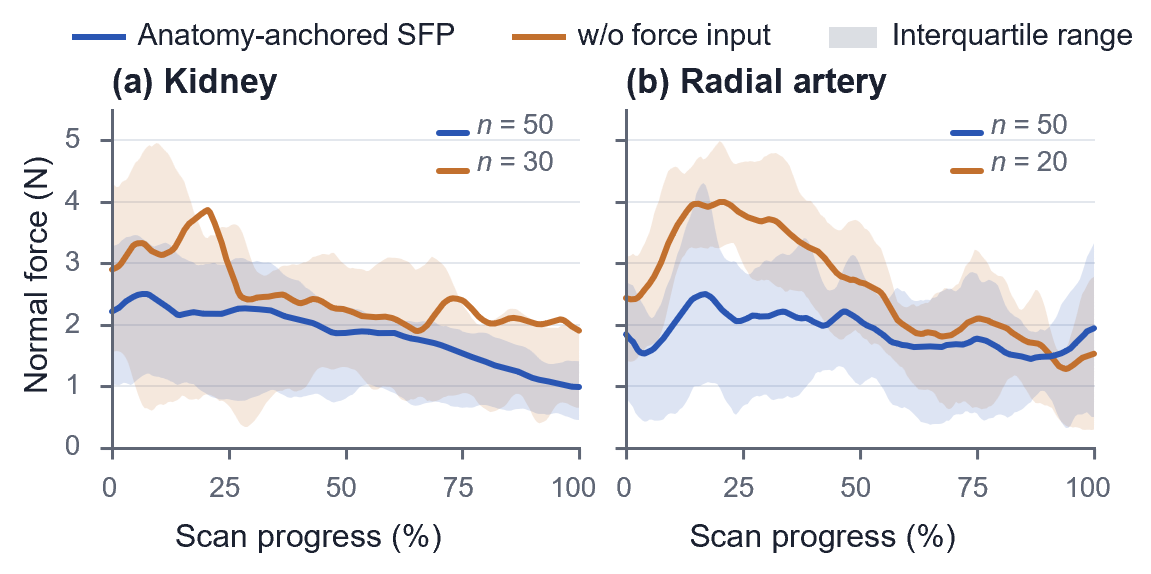}
    \caption{Normal contact force over scan progress for Anatomy-anchored SFP and the \textit{w/o force input} variant in (a) kidney navigation and (b) radial artery sweeping. Lines show the mean and shaded regions the interquartile range across $n$ rollouts.}
    \label{fig:Force_Curve}
\end{figure}

Removing force input reduces task success from $96.0\%$ to $86.7\%$ for kidney navigation and from $94.0\%$ to $90.0\%$ for radial artery sweeping. Its most direct effect, however, appears in the contact behavior. Mean peak force increases from $4.3 \pm 2.0$~N to $5.8 \pm 5.5$~N for kidney navigation and from $4.4 \pm 1.2$~N to $5.4 \pm 1.0$~N for radial artery sweeping, with the much larger standard deviation for kidney navigation indicating occasional rollouts with substantially higher peak forces. The force profiles in Fig.~\ref{fig:Force_Curve} further illustrate this effect: without force input, the normal contact force rises sharply early in the scan and remains higher over much of it, whereas Anatomy-anchored SFP maintains a more uniform profile. 
Compared with the other ablations, the effect on image quality is more moderate. For kidney navigation, the mean ROI score decreases from $3.47$ to $3.07$, the smallest reduction among the kidney variants, and remains above the scale midpoint. For radial artery sweeping, the vessel visibility ratio remains nearly unchanged ($99.6\%$ versus $97.6\%$).
Together, these results suggest that force input provides complementary information about probe-tissue interaction, helping the policy regulate contact forces while maintaining effective navigation.

Unlike anchor-relative poses, which are defined with respect to the initial probe pose and are thus in principle invariant to the setup, absolute poses shift with it. We therefore evaluate the \textit{Absolute pose} variant under both configurations. Even in the nominal configuration, it underperforms the proposed model: for kidney navigation, task success drops from $96.0\%$ to $76.7\%$ and the ROI score from $3.47$ to $2.59$, and for radial artery sweeping it reaches only $75.0\%$ success and $79.1\%$ vessel visibility, compared with $94.0\%$ and $99.6\%$ for Anatomy-anchored SFP even under rotation. Because participants differ in position and body proportions, the absolute pose of the same target varies across episodes even in an aligned setup, which likely hinders learning a consistent motion mapping. Rotating the setup amplifies this effect: kidney success falls further to $70.0\%$, with the lowest ROI score among all kidney variants ($2.27$), and radial artery sweeping fails in all rollouts, likely because the $90^\circ$ rotation redirects the learned long-range sweep away from the vessel. In contrast, Anatomy-anchored SFP is largely unaffected by the rotation, achieving $93.3\%$ success and an ROI score of $3.54$ for kidney navigation and $94.0\%$ success for radial artery sweeping. These results indicate that anchoring probe motion to the anatomical landmark largely removes the dependence on the setup, allowing policies learned from freehand demonstrations to transfer across robot-patient configurations.



\subsection{Full System Validation}
\label{subsec:full_system}
To evaluate the complete pipeline, we integrate anatomy-guided probe placement with the learned navigation policy and conduct end-to-end trials on unseen participants. Unlike the scanning-only experiments with manual probe placement, these trials include external visual observation, anatomical reconstruction, automatic probe placement, and subsequent ultrasound navigation.

Tab.~\ref{tab:full_pipeline} summarizes the results over 30 rollouts per task. The complete system achieves success rates of 100.0\%, 93.3\%, and 90.0\% for kidney, gallbladder, and liver scanning, respectively, and 90.0\% for radial artery sweeping, compared with 96.0\%, 94.0\%, 88.0\%, and 94.0\% in the scanning-only experiments with manual placement. These comparable success rates suggest that automatic placement provides suitable initialization for the learned navigation policies without an apparent reduction in task success.

The peak contact force, however, increases in the complete pipeline for kidney scanning, from $4.3 \pm 2.0$~N with manual placement to $10.2 \pm 6.2$~N. This increase is associated with initial contact over the T12 spinous process during automatic placement. The limited compliance of the tissue overlying this bony prominence leads to a sharp rise in contact force, increasing both the mean and variability of the recorded peak force. For context, the mean peak force remains below the mean maximum forces of 17.4~N and 25.3~N reported for manual abdominal ultrasound examinations in normal- and higher-BMI participants, respectively~\cite{dhyani2017pilot}. A similar increase is observed for radial artery sweeping, from $4.4 \pm 1.2$~N to $8.2 \pm 3.6$~N, which may likewise reflect initial contact at the wrist, where the radial artery lies superficially over the distal radius with little soft tissue to cushion the probe.

The total procedure times are 107.3~s, 116.2~s, and 116.3~s for kidney, gallbladder, and liver scanning, respectively. These durations exceed the scanning-only time-to-target because they also include the initialization stage, in which the robot moves to two observation poses to capture a complete view of the participant lying on the examination bed and the target landmark is localized. HSMR-based anatomical reconstruction alone takes $26.4 \pm 2.2$~s, $24.3 \pm 5.0$~s, and $26.9 \pm 2.0$~s for kidney, gallbladder, and liver scanning, respectively, accounting for roughly one-fifth to one-quarter of the total procedure time. Radial artery sweeping requires a considerably shorter total time of 33.0~s, since the forearm is captured from a single observation pose and wrist localization with MediaPipe Hands requires no HSMR-based skeletal reconstruction.

\begin{table}[t]
\centering
\caption{End-to-end performance of the complete system
(placement and scanning) on unseen participants.}
\label{tab:full_pipeline}

\footnotesize
\renewcommand{\arraystretch}{1.2}
\setlength{\tabcolsep}{0pt}

\begin{tabular*}{\columnwidth}{
    @{\extracolsep{\fill}}lccccc@{}
}
\toprule
\textbf{Task} &
\textbf{Rollouts} &
\shortstack{\textbf{Success}\\(\%) $\uparrow$} &
\shortstack{\textbf{Total time}\\(s) $\downarrow$} &
\shortstack{\textbf{Peak force}\\(N)} &
\shortstack{\textbf{Jerk}\\(m/s$^3$) $\downarrow$} \\
\midrule
Kidney        & 30 & 100.0 & $107.3 \pm 14.6$ 
& $10.2 \pm 6.2$ 
& $0.371 \pm 0.023$ \\
Gallbladder   & 30 & 93.3 & $116.2 \pm 24.1$ 
& $7.0 \pm 3.5$ 
& $0.361 \pm 0.030$ \\
Liver         & 30 & 90.0 & $116.3 \pm 27.1$ 
& $5.0 \pm 1.4$  
& $0.356 \pm 0.056$ \\
Radial artery & 30 & 90.0 & $33.0 \pm 4.3$
& $8.2 \pm 3.6$
& $0.537 \pm 0.054$ \\
\bottomrule
\end{tabular*}

\end{table}

\subsection{Cross-Embodiment Validation}
\label{subsec:cross_embodiment}

\begin{figure}[H]
    \centering
    \includegraphics[width=\linewidth]{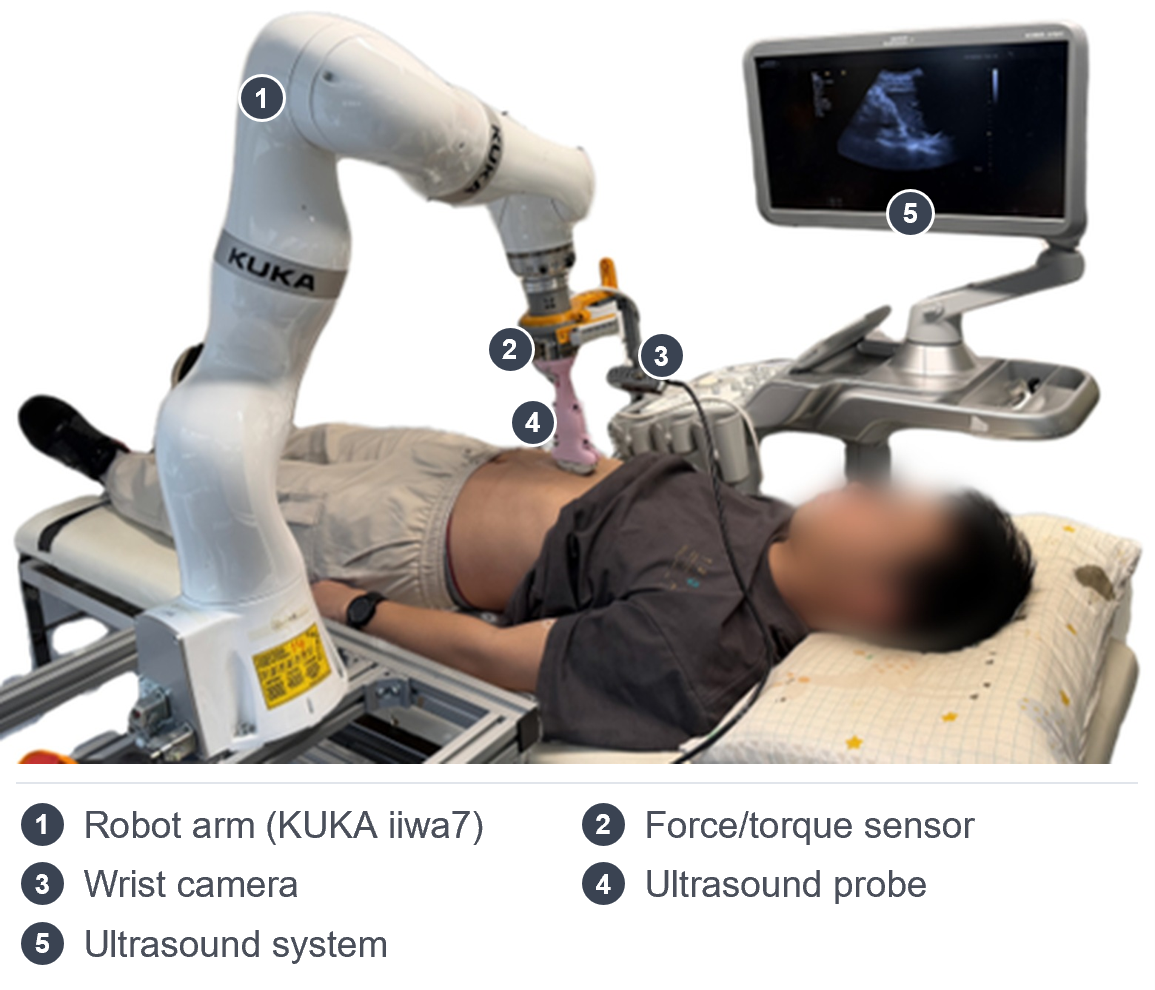}
    \caption{Experimental setup for autonomous robotic ultrasound scanning using KUKA iiwa7.}
    \label{fig:image_scene_kuka}
\end{figure}

To assess whether the learned policies transfer across robotic embodiments, we deploy them on a second robot, a KUKA LBR iiwa 7 R800 (Fig.~\ref{fig:image_scene_kuka}). The same acquisition platform is mounted on the KUKA flange, and the policies trained on the freehand demonstrations are executed without retraining or fine-tuning. Apart from the mechanical mounting and the corresponding end-effector-to-probe transformation, no adaptation is required: the predicted anchor-relative poses are composed with the anchor expressed in the KUKA robot-base frame, as in Eq.~\eqref{eq:relative_to_absolute_pose}. As in Sec.~\ref{subsec:baseline}, the probe is manually positioned at the initial scanning pose. We evaluate all four tasks on five previously unseen participants, with 30 rollouts per task, and summarize the results in Tab.~\ref{tab:cross_embodiment}.

On the KUKA robot, the policies achieve task success rates of $100.0\%$, $86.7\%$, and $93.3\%$ for kidney, gallbladder, and liver scanning, respectively, and $83.3\%$ for radial artery sweeping, compared with $96.0\%$, $94.0\%$, $88.0\%$, and $94.0\%$ on the Franka robot. Averaged over all four tasks, success is thus comparable across the two platforms ($90.8\%$ versus $93.0\%$). Mean peak contact force remains in a similar range ($4.4$-$6.3$~N versus $3.3$-$4.4$~N). Execution on the KUKA robot is, however, slower: time-to-target increases most notably for kidney ($12.7$ versus $5.7$~s), gallbladder ($18.2$ versus $9.7$~s), and radial artery scanning ($24.6$ versus $11.7$~s), and translational jerk is higher for the abdominal tasks ($0.345$-$0.478$ versus $0.217$-$0.292$~m/s$^3$), while remaining comparable for radial artery sweeping. Since the predicted target poses are executed through each robot's own low-level controller, these differences mainly stem from the low-level control and dynamics of the two platforms rather than from the learned policy itself. The lower success rate for radial artery sweeping is attributable to the same cause, as its long continuous trajectory accumulates the effect of these controller differences over more policy updates. Overall, these results show that policies learned from freehand demonstrations can be transferred to a different robotic platform without retraining while maintaining high task success, supporting the embodiment independence of the anchor-relative formulation.

\begin{table}[t]
\centering
\caption{Cross-embodiment validation of autonomous scanning
using a KUKA robot on unseen participants.}
\label{tab:cross_embodiment}

\footnotesize
\renewcommand{\arraystretch}{1.2}
\setlength{\tabcolsep}{0pt}

\begin{tabular*}{\columnwidth}{
    @{\extracolsep{\fill}}lccccc@{}
}
\toprule
\textbf{Task} &
\textbf{Rollouts} &
\shortstack{\textbf{Success}\\(\%) $\uparrow$} &
\shortstack{\textbf{Time-to-Target}\\(s) $\downarrow$} &
\shortstack{\textbf{Peak force}\\(N)} &
\shortstack{\textbf{Jerk}\\(m/s$^3$) $\downarrow$} \\
\midrule
Kidney        & 30 & 100.0 & $12.7 \pm 3.7$ 
& $5.7 \pm 2.8$ 
& $0.431 \pm 0.140$ \\
Gallbladder   & 30 & 86.7 & $18.2 \pm 3.5$ 
& $4.4 \pm 2.0$ 
& $0.345 \pm 0.053$ \\
Liver         & 30 & 93.3 & $10.5 \pm 3.3$ 
& $4.6 \pm 2.5$ 
& $0.478 \pm 0.119$ \\
Radial artery & 30 & 83.3 & $24.6 \pm 5.1$
& $6.3 \pm 3.9$ 
& $0.597 \pm 0.115$ \\
\bottomrule
\end{tabular*}

\end{table}


\section{Conclusion}

This work presented an anatomy-anchored imitation-learning framework that transfers freehand ultrasound scanning skills to autonomous robotic execution. The framework separates scanning into rule-based, anatomy-guided probe placement and learned ultrasound navigation. A Streaming Flow Policy generates probe motion relative to an anchor frame defined at a task-specific anatomical landmark. The policy is conditioned on ultrasound images, wrist camera views, contact force, and anchor-relative probe poses, while a dedicated termination head decides when the target view has been reached. Because probe motion is expressed relative to this anchor rather than in a global frame, demonstrations recorded with the handheld acquisition platform can be used directly for robotic execution, without teleoperation or kinesthetic guidance.

On previously unseen participants, the proposed approach achieved task success rates of $88$-$96\%$ across kidney, gallbladder, and liver standard-plane acquisition and radial artery sweeping. Compared with a fine-tuned $\pi_0$ model, it improved task success by $10$-$42$ percentage points on the abdominal tasks and acquired views of higher physician-rated quality. It also stopped autonomously in more than $93\%$ of successful rollouts, while reducing time-to-target, peak contact force, and trajectory jerk on the abdominal tasks. Ablation studies showed that anatomical anchoring, wrist-camera observation, and force feedback each contribute to performance. For kidney navigation, removing any one of them reduced view quality to approximately the level of the baseline. When the setup was rotated relative to data collection, the anchor-relative representation kept performance largely unchanged, whereas an absolute-pose policy degraded or failed entirely. End-to-end trials further showed that automatic placement provides suitable initialization, with success rates comparable to those obtained with manual placement. Deployed without retraining on a KUKA LBR iiwa 7 R800, the policies achieved $83.3$-$100\%$ success across all four tasks, comparable to the results on the Franka robot.

Building on these results, future work will extend the validation from healthy volunteers to patients with pathology. We further plan to integrate the anatomy-anchored policy as a navigation skill within an agentic framework, such as the LLM-based robotic ultrasound agent of Bi et al.~\cite{bi2026scanning}, enabling guideline-level reasoning and direct patient interaction, such as breath-hold instructions, toward a more interactive and complete autonomous examination.




\bibliographystyle{IEEEtran}
\bibliography{IEEEabrv,references}


 




\vfill

\end{document}